\documentclass[10.5pt,letter]{elsarticle} 
\usepackage[letterpaper, total={6in, 9in}]{geometry}

\usepackage{amsmath,amssymb,amsmath,bm}
\usepackage{color}
\usepackage{times}
\usepackage{url,flushend,multirow,booktabs}
\usepackage[ruled,linesnumbered]{algorithm2e}  % for cross line vertically
\usepackage{bm}
\usepackage{graphicx}
\usepackage{subfigure}
\usepackage{float}
\usepackage{hyphenat}
\usepackage[colorlinks,citecolor=green,urlcolor=blue,bookmarks=false,hypertexnames=true]{hyperref}

\journal{ArXiv}

\usepackage{numcompress}
\newcommand{\eg}{e.g.}
\newcommand{\ie}{i.e.}
\newcommand{\wrt}{\textit{w.r.t.}}
\newcommand{\etal}{\textit{et al.}}
\newcommand{\aka}{\textit{a.k.a.}}

\renewcommand{\bm}{\mathbf}
\DeclareFixedFont{\mf}{OT1}{ptm}{m}{n}{10pt}
\DeclareFixedFont{\mfb}{OT1}{ptm}{bx}{n}{10pt}

\begin{document}

\begin{frontmatter}

\title{Efficient Long-Short Temporal Attention Network for Unsupervised Video Object Segmentation}
%

%% Include affiliations in footnotes:

%\ead[url]{www.elsevier.com}
\author[add1,add2]{Ping~Li}
\author[add1]{Yu~Zhang}
\author[add3]{Li~Yuan}
\author[add4]{Huaxin~Xiao}
\author[add5]{Binbin~Lin\corref{mark1}} 
\cortext[mark1]{Corresponding author}
\ead{binbinlin@zju.edu.cn}
\author[add1]{Xianghua~Xu}

\address[add1]{School of Computer Science and Technology, Hangzhou Dianzi University, Hangzhou, China}
\address[add2]{Guangdong Laboratory of Artificial Intelligence and Digital Economy (SZ), Shenzhen, China}
\address[add3]{School of Electrical and Computer Engineering, Peking University, Beijing, China}
\address[add4]{College of Systems Engineering, National University of Defense Technology, Changsha, China}
\address[add5]{College of Computer Science, Zhejiang University, Hangzhou, China}

\begin{abstract}
   Unsupervised Video Object Segmentation (VOS) aims at identifying the contours of primary foreground objects in videos without any prior knowledge. However, previous methods do not fully use spatial-temporal context and fail to tackle this challenging task in real-time. This motivates us to develop an efficient \emph{L}ong-\emph{S}hort \emph{T}emporal \emph{A}ttention network (termed \textbf{LSTA}) for unsupervised VOS task from a holistic view. Specifically, LSTA consists of two dominant modules, \ie, Long Temporal Memory and Short Temporal Attention. The former captures the long-term global pixel relations of the past frames and the current frame, which models constantly present objects by encoding appearance pattern. Meanwhile, the latter reveals the short-term local pixel relations of one nearby frame and the current frame, which models moving objects by encoding motion pattern. To speedup the inference, the efficient projection and the locality-based sliding window are adopted to achieve nearly linear time complexity for the two light modules, respectively. Extensive empirical studies on several benchmarks have demonstrated promising performances of the proposed method with high efficiency.
\end{abstract}


\end{frontmatter}

%\linenumbers

\section{Introduction}
\label{sec1:intro}

Video Object Segmentation (VOS) task is to localize and segment primary objects in videos, \ie, yielding accurate contours of objects. As a fundamental video processing technique, VOS has found widespread applications, \eg, video editing \cite{sun-pr2020-adaptiveROI}, autonomous driving, and surveillance environment \cite{zhao-pr2021-realtime}, which are highly demanding in real-time processing. Generally, VOS methods are divided into two categories, \ie, \emph{semi-supervised} VOS (\aka, one-shot VOS)  \cite{lan-pr2023-aggregator} which utilizes given object mask of the first frame, and \emph{unsupervised} VOS  (\aka, zero-shot VOS) \cite{lu-cvpr2019-cosnet} for which arbitrary prior knowledge is unavailable during inference. This work concentrates on the more challenging unsupervised VOS, which faces \textbf{two} considerably critical problems: 1) how to find primary objects in video frames; 2) how to speedup object segmentation inference.  % robotics \cite{siam-icra2019-mot},   unsupervised , ren-cvpr2021-rtnet

\begin{figure}[!t]
	\centering
	\includegraphics[width=0.5\textwidth]{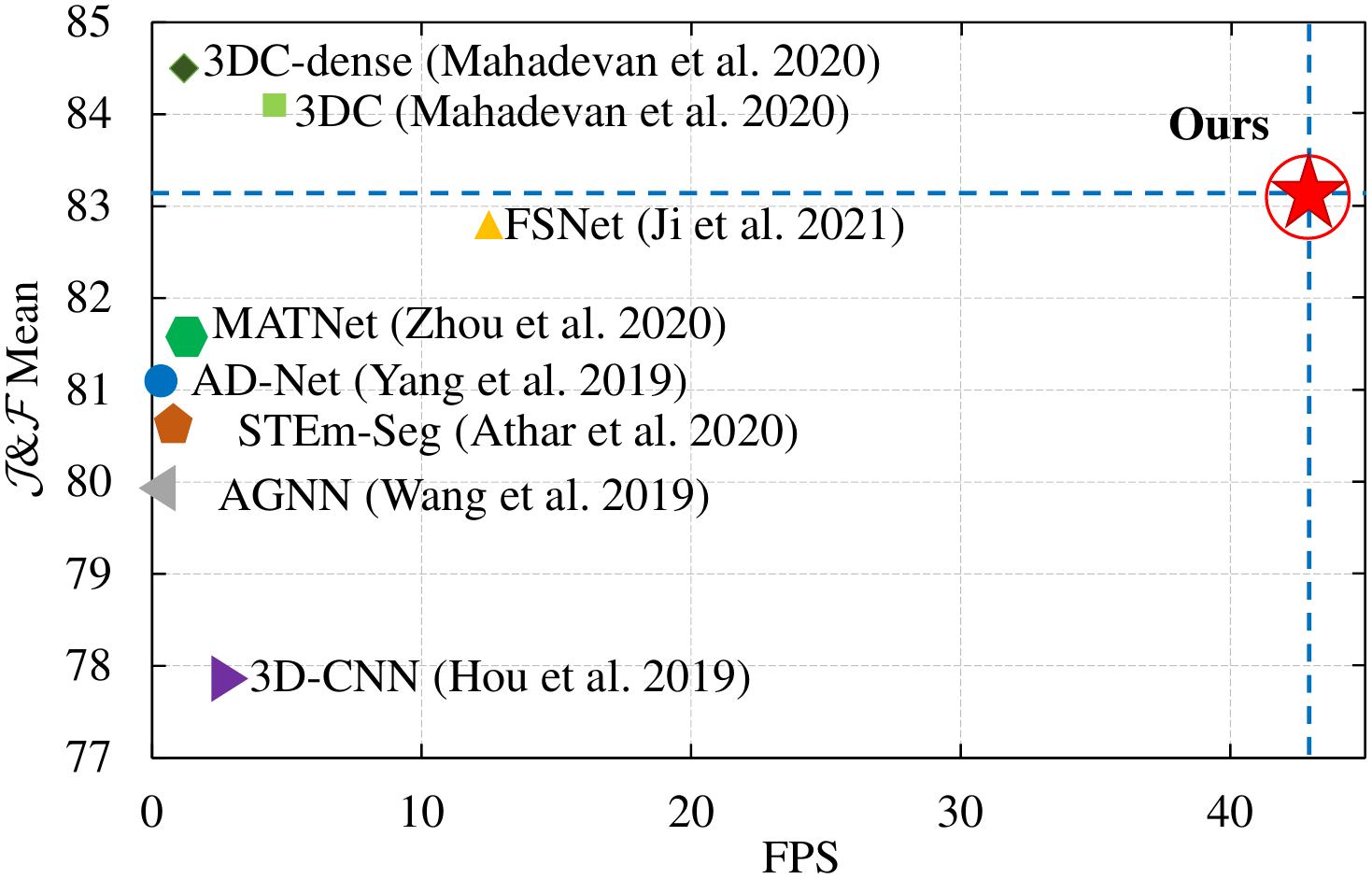}
	\caption{Overall efficiency comparison of several SOTA unsupervised VOS methods without object prior on DAVIS2016 validation set. }
	\label{fig:speed}
\end{figure}

For the first problem, the common insight is to consider salient objects, moving objects, and constantly present objects across video frames. While salient objects attract the visual attention from human eyes, fast moving and drastic deformations may yield objects with small appearance size, leading to less saliency. To model moving objects, someone \cite{ji-iccv2021-fsnet} adopt optical flow technique to capture motion cues, but it is still difficult to discriminate moving objects from dynamic background and usually fails to identify objects in static scenes. From a holistic view, a natural idea is to observe whether there exist constantly present objects in the past frames, and then search objects with similar appearance in the current frame. This idea has been proved effective \cite{lu-cvpr2019-cosnet, wang-iccv2019-agnn} by using dot product attention to encode pixel-wise dense correlations of past frames. However, when partial area of objects are occluded, it adds much difficulty in identifying similar objects due to its strong reliance on appearance. To address these limitations, we model \emph{constantly present objects} and \emph{moving objects} at the same time, by utilizing motion cues and temporal consistency of objects in past frames from both \emph{full-frame} (all pixels in a frame) and \emph{partial-frame} (one frame is separated into many small patches) perspectives. Encoding full-frame pixel correlations facilitates tackling object deformation by modeling appearance pattern, while encoding partial-frame pixel correlations benefits handling object occlusion by modeling pixel movements in the local region of frame.

For the second problem, it still remains an open issue to be explored in unsupervised VOS without any object prior knowledge. Existing models cannot be deployed in real-time applications due to their low inference speed caused by using optical flow \cite{zhou-tip2020-matnet,ji-iccv2021-fsnet} or 3D Convolutional Neural Networks (CNN) \cite{mahadevan-bmvc2020-3dc, athar-eccv2020-stem}, as illustrated in Fig.~\ref{fig:speed}. Accordingly, we explore the way of accelerating inference from both full-frame and partial-frame perspectives to identify objects efficiently. As is well known, the time cost of directly encoding full-frame pixel correlation increases squarely with the number of pixels, which limits its applicability. Someone proposed channel-wise attention \cite{li-eccv2020-gc} to capture the global context of past frames for fast semi-supervised VOS, but it is unable to preserve per-pixel correlations, thus deteriorating performance. Therefore, inspired by the random projection on feature map for computing efficient attention relation with nearly linear complexity \cite{choromanski-iclr2020-performer}, we propose to adopt an efficient projection skill to reveal channel-wise correlation for unsupervised VOS by doing random projection on feature maps derived from CNNs. This projection can achieve the similarity distribution approximation of frames, such that the pixel-wise similarity between the past frames and the current frame can be well preserved in the embedding space. So it is considerably beneficial for discriminating constantly present objects. Meanwhile, since the number of channel $c$ is far less than that of pixel $n$ in a feature map, \ie, $c\ll n$, the time cost of encoding inherent relations among past frames is reduced from square complexity to linear level, \eg, $\mathcal{O}(n^2c) \rightarrow \mathcal{O}(nc^2)$. On the other hand, the locality-based sliding window strategy is employed to partition one full frame to many overlapped patches with size of $k\times k$ ($k\ll n$), \ie, partial frames. This helps to model the local patterns of objects, such as edges, lines, and textures. By this means, encoding partial-frame pixel correlation requires linear time complexity, \ie, $\mathcal{O}(nck^2) = \mathcal{O}(nc) = \mathcal{O}(n)$, much less than directly encoding full-frame correlation, \ie, $\mathcal{O}(n^2c)$. 

Therefore, we propose an end-to-end real-time unsupervised VOS framework, named \textbf{L}ong-\textbf{S}hort \textbf{T}emporal \textbf{A}ttention network (\textbf{LSTA}), to strike a good balance between performance and speed. This framework mainly includes two fast modules, \ie, \emph{Long Temporal Memory} (LTM) and \emph{Short Temporal Attention} (STA). LTM enables encoding long-term full-frame pixel spatiotemporal dependency between the past frames and the current frame, which facilitates identifying constantly present objects. Simultaneously, STA enables capturing short-term partial-frame pixel spatiotemporal relations between one nearby frame and the current frame, which benefits finding moving objects. As a matter of fact, the two modules cooperatively work together to find primary objects by modeling both long-range and short-range spatiotemporal coherence of frames. Meanwhile, it paves the way for discriminating objects from complex background and thus alleviates the object deformation or occlusion problem. More importantly, we apply our proposed efficient projection to LTM and the locality-based sliding window to STA, respectively, for greatly reducing the time complexity. Thus, both LTM and STA can be implemented at linear time complexity, making our LSTA framework very efficient. To examine its performance, we have conducted comprehensive experiments on several benchmark databases, \ie, DAVIS2016\cite{perazzi-cvpr2016-davis16}, DAVIS2017\cite{pont-arxiv2017-davis17}, YouTube-Objects\cite{prest-cvpr2012-yto}, and FBMS\cite{ochs-pami2014-fbms}. Empirical studies demonstrate that our method exhibits promising segmentation performances at a fast speed, \eg, 42.8 fps on 480p resolution videos from DAVIS2016.

Our main contributions are highlighted in the following:
\begin{itemize}
	\item We propose an end-to-end real-time unsupervised VOS framework, called Long-Short Temporal Attention network (LSTA), which enjoys satisfying segmentation accuracy with high inference efficiency.
	
	\item The Long Temporal Memory (LTM) module and the Short Temporal Attention (STA) module are developed to encode both global and local spatiotemporal pixel-wise relations among frames. Hence, constantly present and moving objects can be readily found, and the object deformation or occlusion problem can be alleviated.
	
	\item LTM module and STA module can both achieve the nearly linear time complexity, by respectively adopting the efficient projection and the locality-based sliding window strategy on feature maps.
	
	\item Performance comparisons and extensive ablation studies have justified the realtime segmentation ability with high precision by our method on several benchmarks.
	
\end{itemize}

The rest of this paper is organized as follows. Section~\ref{related} reviews closely related works and Section~\ref{method} introduces the newly developed LSTA framework. After that, we report both quantitative and qualitative experimental results to verify the efficacy of the proposed method in Section~\ref{test}. Finally, we conclude this work in Section~\ref{conclusion}.
%

%-------------------------------------------------------------------------
\section{Related Work}
\label{related}
This section makes a brief summary of closely related VOS methods, including unsupervised, semi-supervised, and fast scenarios. Note that \emph{unsupervised} and \emph{semi-supervised} terms are indicated by whether using the first frame mask during inference. This is a bit different from the traditional machine learning paradigm. For a thorough survey on video segmentation using deep learning techniques, please refer to \cite{zhou-tpami2023-survey}.

\subsection{Unsupervised VOS}
Unsupervised VOS methods have no prior on the first frame mask for inference, making it fairly challenging. Usually, existing methods attempt to find primary objects by considering temporal motion \cite{zhou-tip2020-matnet} or object saliency \cite{song-eccv2018-pdb}. Here, we review two primary kinds of unsupervised methods, \ie, attention-based, and optical flow-based, which adopt 2D convolutions.
% temporal motion \cite{li-cvpr2018-iet}

\textbf{Attention-based} methods \cite{lu-cvpr2019-cosnet}\cite{wang-iccv2019-agnn}\cite{yang-iccv2019-anchor} find objects using appearance feature derived from video frames. For example, COSNet (Co-Attention Siamese Networks) \cite{lu-cvpr2019-cosnet} captures global per-pixel correlation and scene context by using co-attention mechanism on visual features of different frames, which helps to find constantly present objects. But COSNet models spatiotemporal relation between only two nearby frames during inference, which easily causes error accumulation by iterative updates and fails to well capture long-range context of frames. To overcome this drawback, AGNN (Attentive Graph Neural Networks) \cite{wang-iccv2019-agnn} builds fully-connected graph, where a node is the frame feature and an edge stands for the relation of pair-wise features. However, AGNN largely relies on object appearance similarity, and performs poorly when partial objects are occluded. Both COSNet and AGNN utilize dot product attention that requires intensive computations, consequently preventing them from being widely deployed. While attention-based methods focus more on object appearance, partial background areas sharing similar appearance with primary objects will be treated as objects by mistake. To handle this shortcoming, AD-Net (Anchor Diffusion Network) \cite{yang-iccv2019-anchor} adopts instance pruning as a postprocessing to filter out some noisy objects via object detection. In addition, AGS (Attention-Guided object Segmentation) \cite{wang-pami2021-ags} computes visual attention using eye tracking data, and obtains the coarse object location through dynamic visual attention prediction. Nevertheless, AGS employs ConvLSTM (Convolutional Long Short-Term Memory) to model temporal relations, which fails to fully model long-range spatiotemporal context of frames. And a variant of ConvLSTM named RNN-Conv \cite{zhao-pr2021-realtime} aggregates the temporal and the spatial information, such that the model can discover important objects in video.

%\cite{zhuo-tip2020-uovos}\cite{li-cvpr2018-iet}
\textbf{Optical flow-based} methods \cite{zhou-tcsvt2021-fem} capture motion cues from optical flow feature as the compensation of appearance feature. The early work Segflow \cite{cheng-iccv2017-sfl} unifies CNN and optical flow prediction network to predict object mask and optical flow simultaneously, which obtains motion cues in an end-to-end manner; another work \cite{wang-tpami2018-sailency} produces a spatiotemporal edge map by combining static edge probability and optical flow gradient magnitude. But they fail to fully use object appearance features, leading to inferior performance. Thereafter, some works \cite{zhou-tip2020-matnet}\cite{ji-iccv2021-fsnet} concentrate on how to derive and then fuse both the motion feature and the appearance feature. For instance, MATNet (Motion-Attentive Transition Network) \cite{zhou-tip2020-matnet} employs dot product attention to fuse motion and appearance features; RTNet (Reciprocal Transformation Network) \cite{ren_cvpr2021_uvos} mutually evolves the two modalities such that the intra-frame contrast, the motion cues, and temporal coherence of recurring objects are holistically considered; TransportNet \cite{zhang-iccv2021-dtn} employs the Wasserstein distance compute the global optimal flows to transport the features in one modality to the other, and formulates the motion-appearance alignment as an instance of optimal structure matching. But they require heavy computations, and to reduce time cost, FSNet (Full-duplex Strategy Network) \cite{ji-iccv2021-fsnet} makes feature fusion by channel-wise attention, but distills out effective squeezed cues from feature, readily overlooking appearance details. Except for Segflow, the other methods require additional optical flow features, which are not end-to-end and also time-consuming. Besides, optical flow feature mainly encodes the temporal relation between only two nearby frames, failing to model the long-range relation. This usually makes the model perform not well, when drastic changes happen to objects in a long video.
%  which is also used for video super-resolution \cite{song-pr2022-issm}

\subsection{Semi-supervised VOS}
Semi-supervised VOS methods \cite{gao-pr2023-pointbasedmemory,yin-pr2021-agunet} aim to capture objects in video given the first frame mask, which is class-agnostic. Previous methods can be roughly separated into three groups, \ie, \emph{online learning-based}, \emph{attention-based}, \emph{detection-based}.

\textbf{Online learning-based} methods \cite{robinson-cvpr2020-frtm} employ the first frame and its mask to update model parameters during inference, which can adapt to the videos containing various objects in different categories. For example, Sun \etal~\cite{sun-pr2020-adaptiveROI} utilize reinforcement learning to select optimal adaptation areas for each frame, and make the model take optimal actions to adjust the region of interest inferred from the previous frame for online model updating; Lu \etal~\cite{lu-eccv2020-egm} perform the memory updating by storing and recalling target information from the external memory. However, the model updating is very slow, resulting in low segmentation speed.

\textbf{Attention-based} methods \cite{oh-iccv2019-stm} treat the first frame mask as the model input to provide object prior during inference. They encode pairwise pixel dependency among video frames by attention mechanism, which helps to capture objects existing for a long time in the past frames. For example, MUNet (Motion Uncertainty-aware Net) \cite{sun-pr2023-munet} designs a motion-aware spatial attention module to fuse the appearance features and the motion uncertainty-aware feature. The drawback is the high computational cost of computing attention scores, \ie, pairwise pixel similarity.

\textbf{Detection-based} methods \cite{xu-cvpr2019-mhp} always use object detection model to obtain object proposals in each frame, whose feature representations are propagated along the temporal dimension, and generate object masks in line with appearance similarity. The shortcoming is the quality of object proposals will heavily affect the mask quality, and they cannot be trained in an end-to-end way, leading to sub-optimal results.

\subsection{Fast VOS}
Most existing fast VOS methods \cite{robinson-cvpr2020-frtm}\cite{xiao-pami2020-mvos}\cite{wang-cvpr2021-swiftnet}\cite{yin-tnnls2021-ddeal} belong to semi-supervised paradigm, and they strive to efficiently extract discriminant features from video frames. For example, Robinson \etal \cite{robinson-cvpr2020-frtm} propose FRTM (Fast and Robust Target Models) that only updates partial model parameters to speedup inference for online learning-based methods. For attention-based methods, Li \etal \cite{li-eccv2020-gc} use channel-wise similarity to substitute pixel-wise similarity for capturing the global context among frames. Although this substitution can reduce the time complexity, the performance is still far from that of pixel-wise methods such as STM (Space-Time Memory Networks) \cite{oh-iccv2019-stm}. Moreover, Swiftnet \cite{wang-cvpr2021-swiftnet} uses sparse features, \ie, only computing the similarity of those more informative pixels, to reduce dense pixel computations in attention-based methods. 

Real-time unsupervised VOS task still remains less explored in existing works. The one relevant work is WCS-Net (Weighted Correlation Siamese Network) \cite{zhang-eccv2020-wcs} that borrows eye gaze estimation model to provide coarse object location, which is fed into a light segmentation model to obtain object masks of frames. While this approach achieves relatively high inference speed, it does require additional model to yield object prior information as pre-processing step. The other one is Dynamic Atrous Spatial Pyramid Pooling (ASPP) \cite{zhao-pr2021-realtime}, which adopts a dynamic selection mechanism in ASPP, and the dilated convolutional kernels adaptively select appropriate features by the channel attention mechanism. This still requires large computations with an additional RNN-Conv module. Luckily, our LSTA approach can achieve a good balance between segmentation accuracy and inference speed without any object prior, and can be trained in an end-to-end manner.

In addition, referring VOS has recently received more attention from the research field, such as Liang \etal~\cite{liang-tpami2023-locater} explore both local and global temporal context by an improved Transformer model to query the video with the language expression in an efficient manner. Also, the spatio-temporal context is important for weakly-supervised video grounding \cite{li-cvpr2023-winner}, which localizes the aligned visual tube corresponding to a language query. Very recently, Ji \etal~\cite{ji-arxiv2023-segment} explored the Segment Anything Model (SAM) model on a variety of image segmentation tasks, such as agriculture, remote sensing, and healthcare, which may shed some light on the future research of unsupervised VOS task.
\\

% -------------- LSTA-UVOS framework --------------------
\begin{figure*}[!t]
	\centering
	\includegraphics[width=0.9\textwidth]{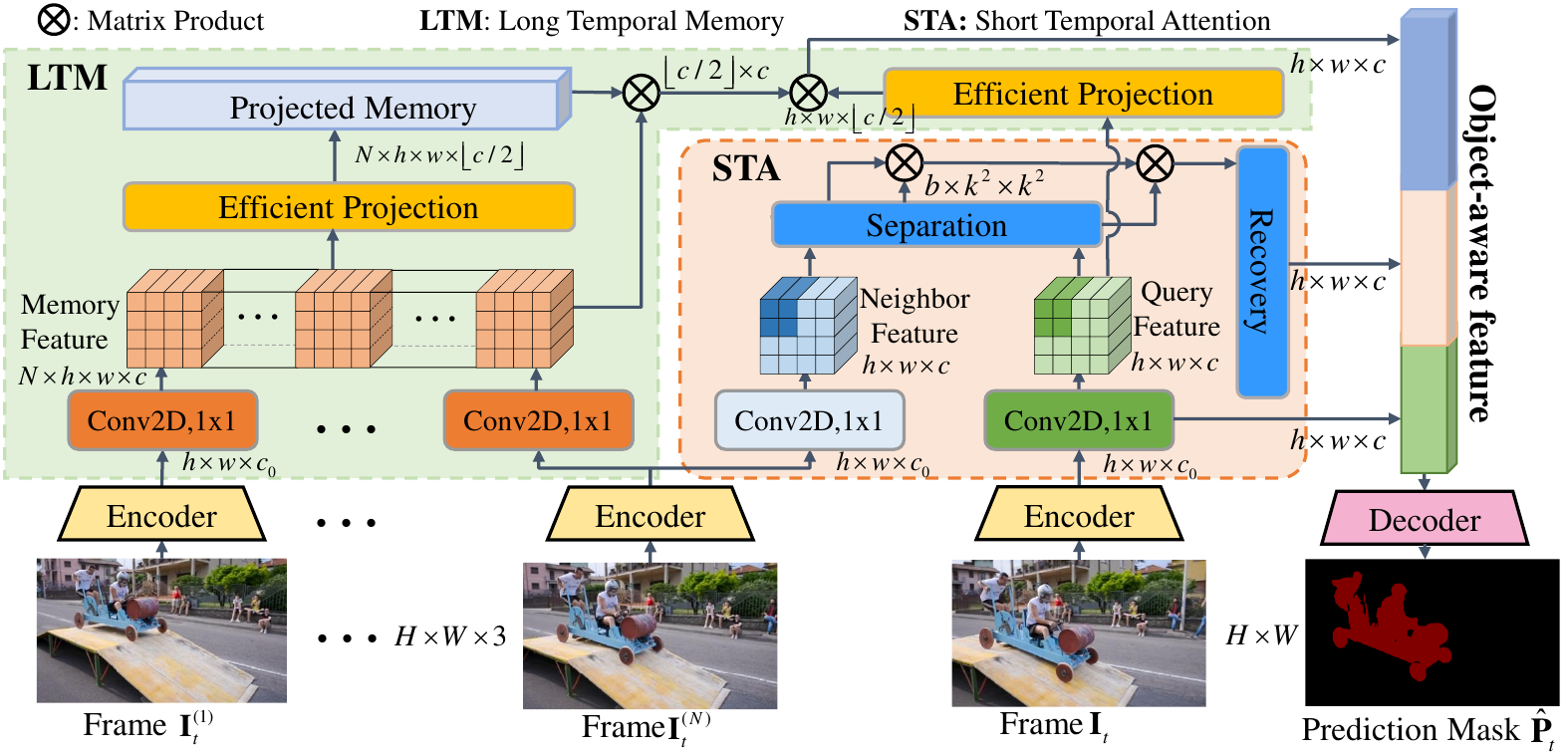}  %MS COCO \cite{lin-eccv2014-coco} 
	\caption{The framework of LSTA model for unsupervised VOS. It is composed of Encoder, LTM, STA, and Decoder. Note that, for the feature map with $c$ channels in LTM and STA, $h$ and $w$ denote height and width, respectively; for STA, $b$ is the number of local patches with size $k$ in one frame after passing the separation layer.}
	\label{fig:LSTA_vos_framework}
	\vspace{-0.5pt}
\end{figure*}

\section{Our LSTA Method}
\label{method}
To efficiently identify primary objects, we develop a real-time end-to-end unsupervised VOS approach, \ie, LSTA, by respecting the spatiotemporal coherence structure in frame data space. First of all, we briefly describe the problem formulation. Then, the main components including Encoder, Long Temporal Memory (LTM) block, Short Temporal Attention (STA) block, and Decoder, in the LSTA framework as illustrated by Fig.~\ref{fig:LSTA_vos_framework}, will be elaborated. 

\subsection{Problem Formulation}
Given a video sequence with $T$ frames, \ie, $\mathcal{V}=\{\mathbf{I}_t\in \mathbb{R}^{H\times W\times 3}|t=1,2, \ldots, T\}$, where $\mathbf{I}_t$ denotes the $t$-th RGB frame with width $W$, height $H$, and three channels. There may exist one or more than one objects in each frame, but no prior knowledge about objects are available. Unsupervised VOS aims to predict the pixel-wise object mask without specifying object in the first frame. For a video sequence $\mathcal{V}$ with one object, the ground-truth mask sequence is $\mathcal{P}=\{\mathbf{P}_t\in \{0, 1\}^{ H\times W}|t=1,2,...,T\}$ and the predicted mask sequence is $\hat{\mathcal{P}}=\{\mathbf{\hat{P}}_t\in\{0, 1\}^{ H\times W}|t=1,2,...,T\}$. The frame mask is a matrix with binary entries, where `0' means background pixel and `1' means primary object pixel. During training, the model uses RGB frames and their ground-truth masks as input, while only RGB frames are available during inference. Note that LSTA does not use all past frames but averagely divides them into $N$ bins, from each of which one frame is randomly selected.  For the current frame $\bm{I}_t$, \ie, query frame, its past frame set is denoted as $\mathcal{I}_t = \{\bm{I}^{(1)}_t, \bm{I}^{(2)}_t, \ldots, \bm{I}^{(N)}_t\}$.

As illustrated in Fig.~\ref{fig:LSTA_vos_framework}, Encoder adopts DeepLab~v3+\cite{chen-eccv2018-deeplabv3} without the last convolution layer, pre-trained on MS COCO database, and it is used to derive object-aware appearance feature from RGB frame $\mathbf{I}\in \mathbb{R}^{H\times W\times3}$, where $H$ denotes height and $W$ denotes width. LTM models long-term full-frame pixel spatiotemporal relation between the past $N$ frames (memory) and the current frame (query) at time step $t$ using channel-wise attention with the efficient projection, which facilitates capturing constantly present objects. STA adopts the locality-based sliding window strategy in the separation layer and attention mechanism on the appearance features of the nearby frame $\mathbf{I}_t^{(N)}$ and the current frame $\mathbf{I}_{t}$, which helps to model the pattern of those moving objects. Decoder that consists of convolution layers, anisotropic convolution block \cite{li-cvpr2020-aic}, and bilinear up-sampling, is used for aggregating features derived from Encoder, LTM, and STA, resulting in object-aware feature representation for computing prediction mask $\hat{\bm{P}_t}\in \mathbb{R}^{H\times W}$ of the $t$-th frame. The recovery layer is used to reshape the feature map with the size of $b\times k^2\times c$ to $h\times w\times c$.

\subsection{Encoder}
To encode the appearance property of video frames, we use the DeepLab v3+ \cite{chen-eccv2018-deeplabv3} model pre-trained on MS COCO database as Encoder, and the last convolution layer is abandoned. As well known, DeepLab v3+ is a typical semantic segmentation model with ResNet101 as its backbone, and the pre-trained model can discriminate a large number of semantic classes of objects.
% DeepLab v3+ \cite{chen-eccv2018-deeplabv3} model pre-trained on MS COCO  \cite{lin-eccv2014-coco}  ResNet101 \cite{he-cvpr2016-resnet} 

All frames in video sequence $\mathcal{V}$ are fed into Encoder to derive its appearance feature map. For the $t$-th frame and its past $N$ frames, we have
\begin{equation}
	\{\mathbf{F}_t^{(1)}, \cdots, \mathbf{F}_t^{(N)}, \mathbf{F}_t \}= 
	           \{\Phi(\mathbf{I}_t^{(1)}), \cdots, \Phi(\mathbf{I}_t^{(N)}), \Phi(\mathbf{I}_t) \}\in \mathbb{R}^{h \times w \times c_0},
\end{equation}
where the function $\Phi(\cdot)$ denotes Encoder which projects RGB frame into feature map $\mathbf{F}_t$ with $c_0$ channels ($c_0$ is 256); the height is $h=\frac{H}{4}$ and the width is $w=\frac{W}{4}$. Since Encoder adopts pre-trained semantic segmentation model, it is able to capture the intrinsic appearance structure of common foreground objects in video frames. This is beneficial for finding those primary objects for segmentation.

% -------------- LTM: Long Temporal Memory block --------------------
\begin{figure}[!t]
	\centering
	\includegraphics[width=0.5\textwidth]{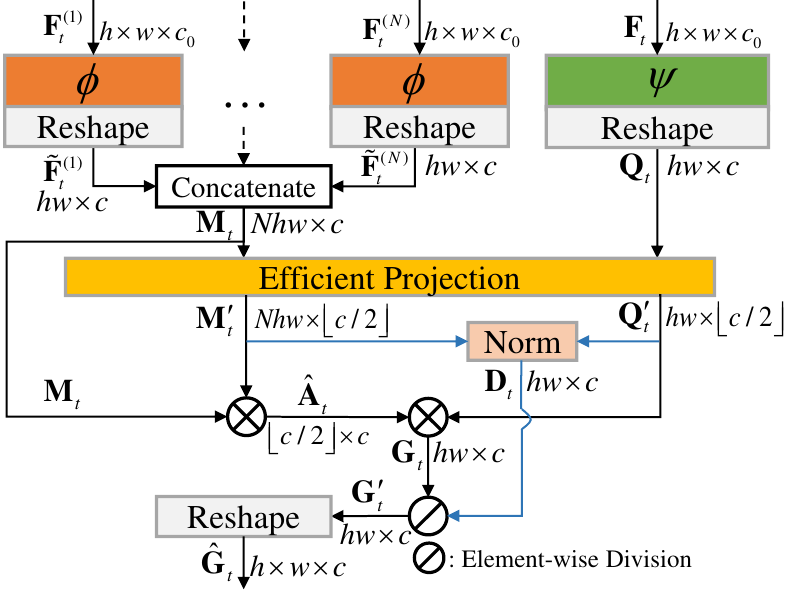}
	\caption{The Long Temporal Memory (LTM) block in LSTA. This block mainly consists of the convolution layer and orthogonal random projection, resulting in projected feature maps. Note that $\phi(\cdot)$ and $\psi(\cdot)$ are convolution layers acting as linear projection. }
	\label{fig:LSTA_LTM}
	\vspace{-0.5pt}
\end{figure}

\subsection{Long Temporal Memory (LTM)}
To identify those constantly present objects in video, LTM block, as illustrated in Fig.~\ref{fig:LSTA_LTM}, employs appearance features of the past frames and the current frame to encode the full-frame pixel spatiotemporal dependency in terms of appearance similarity. This not only helps the model to readily find out those objects with similar appearance in the long-range frame context, such that constantly present objects in the current frame receive more attention, but also makes the model robust to object deformation.

To encode the spatiotemporal relation between the past frames (memory) and the current frame (query), inspired by STM (Space-Time Memory) \cite{oh-iccv2019-stm} for semi-supervised VOS, we use the individual convolution layer to generate the feature map (embedding), which essentially plays the role of key-value maps, so as to reduce the model complexity. The derived feature maps reveal visual semantics for object matching and store detailed cues such as object contours for mask estimation. To determine when-and-where to retrieve related memory feature maps from, we compute similarities between the query feature and the memory features. Query feature is learned to store appearance information for decoding object mask, while memory features are learned to embed visual semantics for object matching. However, densely matching the feature maps of the query and the memory frames, requires expensive computational overheads, \ie, square time complexity \wrt~the number of pixels. This motivates us to model channel-wise correlation rather than pixel-wise one, and the cost is greatly reduced by using smaller channel number. However, channel-wise attention may break down the pixel-wise similarity distribution (\eg, probability histogram), since all the pixels of pair-wise memory feature maps are taken into account channel by channel rather than pixel by pixel. Hence, we propose to make an efficient projection on the pixel-wise feature embedding of the past frames and the current frame using the similar projection skill in \cite{choromanski-iclr2020-performer}. 

As illustrated in Fig.~3, the input of LTM block is the appearance feature map set $\mathcal{F}_t = \{\bm{F}^{(1)}_t, \bm{F}^{(2)}_t, \ldots, \bm{F}^{(N)}_t, \bm{F}_t\}\in \mathbb{R}^{h\times w\times c_0}$ of $N$ past frames and the current frame $\bm{I}_t$. The feature maps of past frames and that of current frame are respectively fed into two 2D convolution layers $\phi(\cdot)$ and $\psi(\cdot)$ with $1\times 1$ kernel, followed by reshaping the height and the width dimensions, \ie, $h\times w \rightarrow hw = n$. This results in memory features $\{\tilde{\bm{F}}^{(s)}_t \in \mathbb{R}^{hw\times c}\}_{s=1}^N$  ($c$ is 128) and query feature $\bm{Q}_t\in \mathbb{R}^{hw\times c}$, where memory features are concatenated along row dimension into one matrix, called feature memory $\mathbf{M}_t$, \ie,
\begin{equation}
	\mathbf{M}_t= [\tilde{\bm{F}}^{(1)}_t, \ldots, \tilde{\bm{F}}^{(N)}_t] \in \mathbb{R}^{Nhw\times c},
\end{equation}
where $[\cdot, \cdot]$ denotes the concatenation, $N\times hw = Nhw$.

To preserve the pixel-wise similarity distribution, LTM conducts random projection on feature memory $\bm{M}_t$ and query feature $\bm{Q}_t$ at pixel level, leading to projected pixel values, \ie,
\begin{align}\label{eq:project_pixel}
	m^{\prime} & = \frac{1}{\sqrt{\lfloor c/2 \rfloor }}\exp(\mathbf{u}^T\mathbf{m}- \frac{||\mathbf{m}||_2^2}{2}), \\
	q^{\prime}  & = \frac{1}{\sqrt{\lfloor c/2 \rfloor }}\exp(\mathbf{u}^T\mathbf{q}- \frac{||\mathbf{q}||_2^2}{2}),
\end{align}
where the pixel feature vector $\bm{m}\in \mathbb{R}^c$ is stacked in each row of feature memory matrix $\bm{M}_t$, and the pixel feature vector $\bm{q}\in \mathbb{R}^c$ is stacked in each row of query feature matrix $\bm{Q}_t$;  the vector $\bm{u} \in \mathbb{R}^c$ is an orthogonal projection vector, which is randomly initialized for each projection; the constant $\lfloor c/2 \rfloor$ is a scaling factor and $\lfloor \cdot \rfloor$ rounds down fractions. Thus, we can obtain the projected pixel feature vectors $\bm{m}^{\prime}\in \mathbb{R}^{\lfloor c/2 \rfloor}$ and $\bm{q}^{\prime}\in \mathbb{R}^{\lfloor c/2 \rfloor}$ for each pixel in the memory feature map and the query feature map, respectively, by doing orthogonal random projections for $\lfloor c/2 \rfloor$ times as in (\ref{eq:project_pixel}). All projected pixel feature vectors are collected together to be reshaped into the matrix with the same size of that of unprojected feature matrix, \ie, $\mathbf{M}^\prime_t \in \mathbb{R}^{Nhw\times \lfloor c/2 \rfloor}$ and $\bm{Q}^\prime_t\in \mathbb{R}^{hw\times \lfloor c/2 \rfloor}$.

Therefore, the pixel-wise appearance similarity between the past frames and the query frame can be revealed by the product of projected pixel feature matrices $\mathbf{M}^\prime_t$ and  $\mathbf{Q}^\prime_t$, \ie, $\mathbf{A}_t=  \mathbf{Q}^\prime_t \mathbf{M}^{\prime \top}_t \in \mathbb{R}^{hw\times Nhw}$,
where the large values taken by the entries of appearance similarity matrix $\mathbf{A}_t$ indicate that the current frame shares higher similarity with the past frames in appearance. Meanwhile, its elements can be treated as attention weights of memory frames. Thus, we obtain the global feature representation by $\mathbf{G}_t=  \mathbf{A}_t \mathbf{M}_t \in \mathbb{R}^{hw\times c}$, which provides the guidance to retrieve the relevant memory frames with highly similar appearance, to attend on the query frame.

Usually, the same object constantly present in video will share common appearance across frames, so learning global feature representation contributes to locating primary objects. However, the above process requires high time complexity, \ie, $\mathcal{O}(n^2 c)$,  for obtaining
\begin{equation} \label{eq:G_origin}
	\mathbf{G}_t = \mathbf{Q}^\prime_t \mathbf{M}^{\prime \top}_t \mathbf{M}_t  \in \mathbb{R}^{hw\times c}.
\end{equation}

To make the model more efficient, we first compute the channel-wise memory similarity matrix, \ie,
\begin{equation} \label{eq:A_hat}
	\hat{\mathbf{A}}_t=  \mathbf{M}^{\prime \top}_t \mathbf{M}_t \in \mathbb{R}^{\lfloor c/2 \rfloor \times c},
\end{equation}
where $\mathbf{M}_t \in \mathbb{R}^{Nhw \times c}$ and $\mathbf{M}^\prime_t \in \mathbb{R}^{Nhw\times \lfloor c/2 \rfloor}$, and models the channel-wise correlation of memory features. Then, we calculate the cheap global feature representation by substituting Eq.~(\ref{eq:A_hat}) into Eq.~(\ref{eq:G_origin}), resulting in
\begin{equation}
	\mathbf{G}_t = \mathbf{Q}^\prime_t \hat{\mathbf{A}}_t \in \mathbb{R}^{hw\times c},
\end{equation}
which preserves the pixel-wise similarity distribution of the feature memory and the query feature. Especially, this formulation only requires $\mathcal{O}(n c^2)$, which can be further simplified to $\mathcal{O}(n)$ when $c\ll n$. In another word, the time complexity is linear with the number of feature map pixels, allowing the model to run very efficiently.

Besides, to avoid unstable numerical solutions due to large values, we do normalization
on $\mathbf{G}_t$, \ie,
\begin{equation}
	\mathbf{G}^\prime_{t}= {\mathbf{G}}_t   \oslash  \mathbf{D}_t \in \mathbb{R}^{hw\times c},
\end{equation}
where the symbol $ \oslash $ denotes element-wise division (Hadamard division), and the matrix $\mathbf{D}_t = \mathbf{Q}^\prime_t\cdot (\mathbf{M}^{\prime\top}_t \mathbf{1})\in \mathbb{R}^{hw\times c}$ is used for normalization. Here, $\bm{1}$ is an all-one matrix with the size of $Nhw\times c$. In the end, we reshape the matrix $\mathbf{G}^\prime_{t}$ to $\hat{\mathbf{G}}_{t}\in\mathbb{R}^{h\times w \times c}$, \ie, the global feature map, since it reflects the long-range inherent semantic relations of memory frames and the query frame.

% -------------- STA: Short Temporal Attention block --------------------
\begin{figure}[!t]
	\centering
	\includegraphics[width=0.50\textwidth]{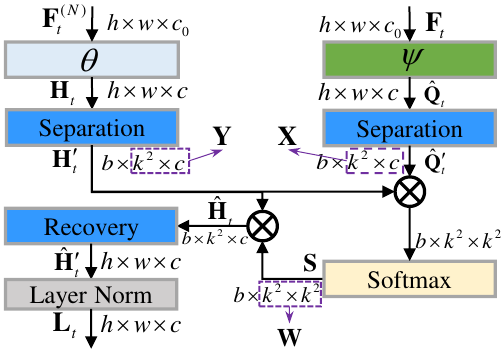}
	\caption{The Short Temporal Attention (STA) block in LSTA. This block mainly consists of the 2D convolution layer with $1\times 1$ kernel, separation and recovery operations. }
	\label{fig:LSTA_STA}
	\vspace{-0.5pt}
\end{figure}

\subsection{Short Temporal Attention (STA)}
To identify moving objects, STA block as illustrated in Fig.~\ref{fig:LSTA_STA}, encodes the partial-frame pixel spatiotemporal dependency of the nearest past frame and the current frame, by discovering the object motion pattern in terms of local patches of the appearance feature map. This not only helps to capture moving objects in short-term frame context, but also makes the model robust to occlusions in video.

Motivated by the attention mechanism \cite{vaswani-nips2017-attention} and the fact that only partial pixels in neighboring frames will change, we propose a patch-based technique called STA, for encoding temporal attention locally. Unlike the vanilla attention modeling global relations of all pixel pairs in full frame, STA models local relations of limited pixel pairs in partial frame sequentially. This is achieved by adopting the locality-based sliding window strategy, which separates one frame into a number of much smaller regions called patches. Then, STA models the local spatiotemporal relation of pixel patches between the current frame and the nearest past frame. 

As illustrated in Fig.~\ref{fig:LSTA_STA}, the inputs of STA are the appearance feature maps of the current frame $\bm{I}_t$ and the nearest past frame $\bm{I}^{(N)}_t$, \ie, $\bm{F}_t$ and $\bm{F}^{(N)}_t$. At first, the channel dimension of $\bm{F}^{(N)}_t$ is reduced to $c$ by a $1\times 1$ convolution $\theta(\cdot)$, leading to the neighbor feature map $\mathbf{H}_{t} = \theta(\mathbf{F}_{t}^{(N)})\in \mathbb{R}^{ h \times w \times c}$. For $\bm{F}_t$, we directly utilize its feature matrix $\mathbf{Q}_t\in\mathbb{R}^{hw\times c}$ from the convolution layer $\psi(\cdot)$ in LTM, and reshape it to query feature map $\hat{\mathbf{Q}}_t \in \mathbb{R}^{ h \times w \times c}$.

\begin{figure}[!t]
	\centering
	\includegraphics[width=0.6\textwidth]{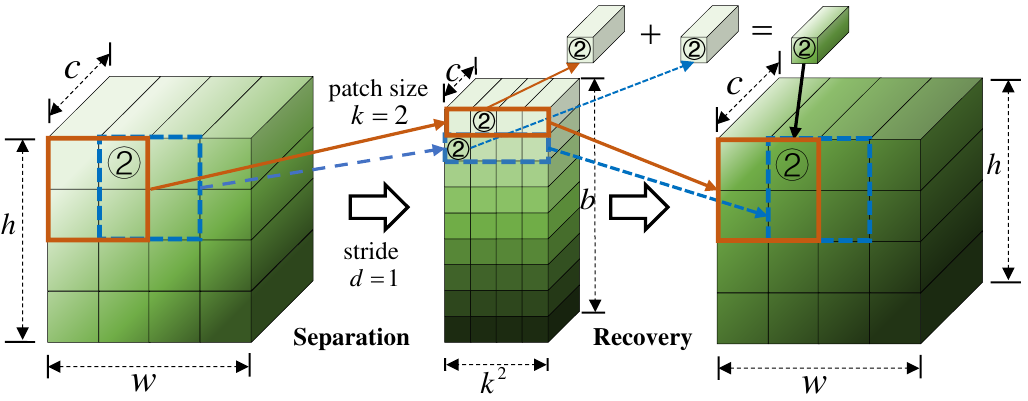}
	\caption{The illustration of separation and recovery layers in STA block.}
	\label{fig:LSTA_divide}
	\vspace{-0.5pt}
\end{figure}

STA adopts the locality-based sliding window strategy, which makes the model cheap to learn. Assume the patch size is $k$ and each feature map is separated into $b$ patches, as indicated by Fig.~\ref{fig:LSTA_divide}, we have $n=hw=h\times w= k\times k\times b$ pixels, and the time complexity is $\mathcal{O}(nk^2c) = \mathcal{O}(nc) = \mathcal{O}(n)$ ($k^2\ll n$, $c\ll n$). Here we neglect the stride factor, as it does not change time complexity compared to the number of pixels. STA models the spatial correlation of each patch with $k\times k\times c$ pixels in a feature map, and the stride $1\le d < k$ affects the number of patches, \ie, $b=(h-k+1)/d\times (w-k+1)/d$ with zero padding. Here, $k=8$ and $d=4$. As a result, we can obtain query patch feature tensor $\hat{\mathbf{Q}}^\prime_t \in \mathbb{R}^{b\times k^2 \times c}$ and neighbor patch feature tensor $\mathbf{H}^\prime_t \in \mathbb{R}^{b \times k^2 \times c}$, which are composed of $b$ patch matrices, \ie,
$\{\mathbf{X}_1,\mathbf{X}_2, \ldots, \mathbf{X}_b\}\in \mathbb{R}^{k^2 \times c}$ and $\{\mathbf{Y}_1,\mathbf{Y}_2, \ldots, \mathbf{Y}_b\}\in \mathbb{R}^{k^2 \times c}$, acting as tensor slices. 

% -------------- STA: patch separation and  recovery --------------------

To discover the pixel moving pattern in the local region of frame, STA computes the semantic similarity of query-neighbor patch pair, \ie, $(\bm{x}_i, \bm{y}_j)$, where $\bm{x}_i \in \mathbb{R}^c$ is stacked in the $i$-th row of $\bm{X}$ and $\bm{y}_j \in \mathbb{R}^c$ is stacked in the $j$-th row of $\bm{Y}$. Then, we obtain similarity value of each pixel pair by
\begin{equation}
	\bm{w}_{ij} =\frac{\exp(\mathbf{x}_i^\top \mathbf{y}_j / \sqrt{c})}{\sum_j^{k^2} \exp(\mathbf{x}_i^\top \mathbf{y}_j /\sqrt{c})}.
\end{equation}

In this way, we can compute the similarity patch by patch, and thus get the semantic similarity tensor $\mathbf{S}\in \mathbb{R}^{b\times k^2\times k^2}$, consisting of $b$ matrices $\{\mathbf{W}_1,\mathbf{W}_2, \ldots, \mathbf{W}_b\}\in \mathbb{R}^{k^2 \times k^2}$. They help to capture the local semantic coherence of feature pairs. For those moving objects, their semantic similarity can be well encoded in the short-term spatiotemporal context, by revealing the hidden pattern. Actually, the similarity matrices play an important role in retrieving dynamics of moving objects from neighbor frame at patch level, \ie, $\{\bm{W}_1 \bm{Y}_1, \ldots, \bm{W}_b \bm{Y}_b\}\in \mathbb{R}^{k^2\times c}$, which further act as slices of the tensor $\hat{\mathbf{H}}_{t}\in \mathbb{R}^{b\times k^2\times c}$, \ie, the local feature representation.

To preserve the spatial pixel correlations, we reshape $\hat{\mathbf{H}}_{t}$ to the feature map $\hat{\mathbf{H}}^\prime_{t} \in \mathbb{R}^{h\times w\times c}$ via the recovery layer, which is essentially the inverse process of patch separation. For each slice $k^2\times c$, the $k^2$ entries in every column are reshaped to a local patch with the size of $k\times k$, in which way the entries in one column of all $b$ slices are reshaped to a matrix with the size of $h\times w$. Then, adding the channel dimension $c$ leads to the recovered feature map $h\times w\times c$. Due to the stride of sliding window, there is redundant samplings on feature map. Hence, we simply sum the elements of those local patch features which are projected onto the same pixel, resulting in the feature map with the same size of the original one. However, those large sum values might possibly lead to numerical instability of the data distribution. To deal with this issue, we impose the layer normalization on $\hat{\mathbf{H}}^\prime_{t}$, \ie, normalization across all channels of feature map. As a result, we obtain the local feature map $\mathbf{L}_{t}\in \mathbb{R}^{h \times w \times c}$, which encodes the short-range spatiotemporal pixel relations of the nearest neighbor frame and the query frame in terms of small patches. 

\subsection{Decoder}
Decoder is composed of convolution layer, anisotropic convolution (AIC) \cite{li-cvpr2020-aic} block (consisting of several 2D convolution layers), and up-sampling operation. Its goal is to consider both the long-range and the short-range spatiotemporal pixel correlations of the past frames and the current frame, by make a fusion on three data streams, \ie, global feature map $\hat{\mathbf{G}}_{t}$, local feature map $\mathbf{L}_{t}$, and query feature map $\hat{\mathbf{Q}}_t$.

Particularly, we first concatenate the above three feature maps along the channel dimension into a unified feature map with size of $h\times w\times 3c$, whose channel dimension is reduced to $c$ via one $3\times 3$ 2D convolution layer. Then, the unified feature map is fed into the AIC module that helps to discriminate objects from the spatial context. This is followed by passing the other $3\times 3$ 2D convolution layer for reducing the channel dimension into 2. Hereafter, the resolution of feature map is enlarged to that of original RGB frame by bilinear up-sampling. Finally, it yields the object-like pixel probability using softmax function, namely
\begin{equation} \label{eq:objectpixelprob}
	\tilde{\mathbf{P}}_t =\Theta(\hat{\mathbf{G}}_{t},\mathbf{L}_{t},\hat{\mathbf{Q}}_t)\in \mathbb{R}^{ H\times W \times 2},
\end{equation}
where $\Theta(\cdot)$ denotes Decoder. The elements of the first (index 0) and the second (index 1) channel denote the probabilities of pixels belonging to background and primary object, respectively. This probability tensor can be easily transformed to a binary matrix by taking the channel index of the higher probability for each pixel, \ie, $\hat{\mathbf{P}}_t = \{0, 1\} \in \mathbb{R}^{H\times W}$.

\subsection{Loss Function}
To optimize the LSTA model, we adopt the Cross-Entropy (CE) loss with the online hard example mining strategy \cite{shrivastava-cvpr2016-ohem}. It selects those hard pixels (usually those with larger loss values) to calculate the CE loss, which is beneficial for promoting the robustness of discriminating those ambiguous pixel regions. The model loss is computed by
\begin{equation} \label{eq:loss1}
	\mathcal{L}_1(\tilde{\mathbf{P}}_t ,\mathbf{P}_t)= \overline {\text{Max}}_r ( \left \{ -p^0_{tz} \log \tilde{p}^0_{tz} - p^1_{tz} \log \tilde{p}^1_{tz} \right \}_{z=1}^{HW}),
\end{equation}
where $ \overline {\text{Max}}_r (\cdot)$ ($r = \left\lfloor \frac{HW}{16} \right\rfloor $ and \scalebox{0.8}{$HW= H\times W$}) means taking the average of those $r$ largest loss values of pixels, $\{\tilde {p}_{tz}^0, \tilde {p}_{tz}^1\}$ are the predicted probability values of the $z$-th pixel in current frame $\bm{I}_t$, and $\{p_{tz}^0, p_{tz}^1\}$ are the corresponding ground-truth object mask values.

Inspired by knowledge distillation, we utilize the semi-supervised VOS model, \ie, STM \cite{oh-iccv2019-stm}, as teacher network, which provides guidance for our unsupervised VOS model, \ie, LSTA, as student network. Note that, the STM model trained on DAVIS17 \cite{pont-arxiv2017-davis17} and YouTube-VOS \cite{xu-arxiv2018-ytbvos} does not participate in training our LSTA model but only helps to yield the initial pixel probability of frames, \ie, soft labels, to guide the loss computation. And the inference process does not involve teacher network as well. Assume that the initial pixel probability from teacher network is $\bar{\mathbf{P}}\in\mathbb{R}^{ H\times W \times 2}$, we compute the following loss:
\begin{equation} \label{eq:loss2}
	\mathcal{L}_2(\tilde{\mathbf{P}}_t, \bar{\mathbf{P}}_t)= \frac{1}{HW}\sum_{z=1}^{HW}( -\bar{p}^0_{tz} \log \tilde{p}^0_{tz} - \bar{p}^1_{tz} \log \tilde{p}^1_{tz}),
\end{equation}
where $\{\bar{p}_{tz}^0, \bar{p}_{tz}^1\}$ are the initial probability values of the $z$-th pixel in the current frame $\bm{I}_t$.

Therefore, the total loss of our LSTA model is
\begin{equation} \label{eq:loss}
	\mathcal{L}(\tilde{\mathbf{P}}_t ,\mathbf{P}_t, \bar{\mathbf{P}}_t) = \alpha \mathcal{L}_1(\tilde{\mathbf{P}}_t ,\mathbf{P}_t) + (1-\alpha)\mathcal{L}_2(\tilde{\mathbf{P}}_t, \bar{\mathbf{P}}_t),
\end{equation}
where the tradeoff parameter $\alpha>0$ is used to balance the contribution of the two loss terms to the objective function. Here, we use an empirical value 0.5.

To summarize our approach, we briefly show the training process in Algorithm~\ref{alg:train} and the inference process in Algorithm~\ref{alg:inference}. Our method is relevant to the widely used LSTM and Transformer, which captures the short-term  and the long-term temporal relations, respectively. However, their working mechanisms are different. In particular, LSTM adopts the gating skill to control the history information in a sequence, while our STA block uses the attention mechanism to discover the object motion pattern in terms of local patches of the appearance feature map, which encodes the partial-frame pixel spatiotemporal dependency with the locality-based sliding window strategy. Moreover, Transformer adopts the self-attention to model the global dependency and needs the square time complexity, while our LTM block encodes the full-frame pixel spatiotemporal dependency in terms of appearance similarity at the cost of almost linear complexity by computing the channel-wise memory similarity matrix.

% ------------------------- LSTA-UVOS Training algorithm -------------------
\begin{algorithm}[!t]
	\caption{Training Process of LSTA model.}
	\label{alg:train}
	\KwIn{Training videos; ground truth mask $\mathcal{P}$; model parameters $\Omega$; number of past frames $N$; $\alpha=0.5$; $Iter_{max}=7.5e4$. }
    Randomly select $T$ frames as query for each video. \\
	\While{not converged}{
    Randomly select one query frame $\bm{I}$ from one video in sequence. \\
    Select $N$ past frames $\mathbf{I}^{(1)},\mathbf{I}^{(2)},...,\mathbf{I}^{(N)}$ . \\
	Feed query frame and past frames into Encoder to obtain appearance feature map $\mathbf{F}$ and $\{ \mathbf{F}^{(1)},\mathbf{F}^{(2)}, \ldots, \mathbf{F}^{(N)}\} $. \\
	Input appearance feature maps $\{ \mathbf{F}, \mathbf{F}^{(1)},\mathbf{F}^{(2)},...,\mathbf{F}^{(N)} \}$ into LTM to obtain global feature map $\hat{\mathbf{G}}$.\\
	Feed appearance feature maps $\mathbf{F}$ and $\mathbf{F}^{(N)} $ into STA to obtain local feature map $\mathbf{L}$ and query feature map $\hat{\mathbf{Q}}$.\\
	Obtain the object-like pixel probability $\tilde{\mathbf{P}}$ using $\hat{\mathbf{G}}$, $\mathbf{L}$ and $\hat{\mathbf{Q}}_t$ as in Eq.~(\ref{eq:objectpixelprob}). \\
    Calculate the first loss $\mathcal{L}_1$ as in Eq.~(\ref{eq:loss1}) and the second loss $\mathcal{L}_2$ as in Eq.~(\ref{eq:loss2}). \\
	Calculate the total loss function $\mathcal{L}$ in Eq.~(\ref{eq:loss}). \\
	Update model parameters $\Omega$ using SGD. \\
	}
	\KwOut{Trained model.}
\end{algorithm}
%

% ------------------------- LSTA-UVOS Inference algorithm -------------------
\begin{algorithm}[!t]
	\caption{Inference Process of LSTA model.}
	\label{alg:inference}
	\KwIn{Video frames; trained model.}

    Obtain appearance feature map for every video frame by Encoder. \\
    Feed appearance feature maps of query frame and past frames to LTM to obtain global feature map. \\
    Feed appearance feature maps of query frame and nearest neighbor frame to STA to obtain local feature map and query feature map. \\
    Input global feature map, local feature map, and query feature map to Decoder, which yields object mask. \\

    \KwOut{Object mask set.}
\end{algorithm}

\section{Experiments}
\label{test}
This section shows comprehensive empirical studies on several benchmark data sets. All experiments were conducted on a machine with NVIDIA Titan RTX Graphics Card for training and NVIDIA Titan Xp for inference, and our LSTA model was compiled using PyTorch 1.8, Python 3.6, and CUDA 11.0.

%-------------------------------------------------------------------------
\subsection{Data Sets}
In total, there are four publicly available VOS data sets used in the experiments. Details are shown below.

\textbf{DAVIS}$\footnote{https://davischallenge.org/index.html}$ short for Densely Annotated Video Segmentation, provides two kinds of frame resolution, \ie, 480p and 1080p, with pixel-level frame mask. It has two versions, \ie, DAVIS2016\cite{perazzi-cvpr2016-davis16} and DAVIS2017\cite{pont-arxiv2017-davis17}, involving various scenes, such as animal, sports, and traffic vehicles. The former has 50 video sequences, which are divided into 30 training videos and 20 validation videos for inference; each video contains only one object, and there are 3,455 frames with ground-truth masks. The latter DAVIS2017 is an expansion of the former, and the number of videos increases to 150, among which there are 90 videos (60 for training and 30 for validation) with 10,459 frames with ground-truth masks; each video may contain more than one objects, adding difficulty to the task, and there are totally 376 objects.

\textbf{YouTube-VOS}$\footnote{https://youtube-vos.org/dataset/vos/}$ \cite{prest-cvpr2012-yto} collects video clips from YouTube web site, including various classes, such as animal, transportation, accessory, and human event. Each clip usually contains multiple objects, with a duration of 3s to 6s. It has three subsets, and we only use its training set, including 3,471 videos with dense (6 fps) object annotations, 65 categories, and 5,945 unique object instances.

\textbf{YouTube-Objects}$\footnote{https://data.vision.ee.ethz.ch/cvl/youtube-objects/}$ \cite{prest-cvpr2012-yto} is composed of 126 videos collected from YouTube by querying for the names of 10 object classes. The duration of each video varies between 30s and 180s, and each video contains one object. Following \cite{zhou-tip2020-matnet}\cite{wang-iccv2019-agnn}, we use all videos for inference.

\textbf{FBMS}$\footnote{https://lmb.informatik.uni-freiburg.de/resources/datasets/moseg.en.html}$ \cite{ochs-pami2014-fbms} short for Freiburg-Berkeley Motion Segmentation, contains 59 video sequences, which are separated to 29 training videos and 30 validation videos, each of which has one object. There are 720 frames with pixel-level mask annotations, which are made with an interval of 20 frames. Following \cite{zhou-tip2020-matnet}\cite{lu-cvpr2019-cosnet}, we only use validation set for inference.

\subsection{Evaluation Metrics}
We evaluate our LSTA model on DAVIS, YouTube-Objects, and FBMS benchmarks, and the model is learned using the training sets of DAVIS2017 and YouTube-VOS. Following the previous works \cite{zhou-tip2020-matnet}\cite{lu-cvpr2019-cosnet}\cite{yang-iccv2019-anchor}, we use region similarity $\mathcal{J}$, contour accuracy $\mathcal{F}$, and $\mathcal{J}\&\mathcal{F}$ as the evaluation criteria. For DAVIS, we used the official benchmark code \cite{perazzi-cvpr2016-davis16}. For YouTube-Objects and FBMS, we use $\mathcal{J}$ Mean as the metric.

Region similarity $\mathcal{J}$ is the Intersection over Union (IoU, namely Jaccard coefficient) of predicted mask $\hat{\mathbf{P}}$ and ground-truth mask $\mathbf{P}$, which reflects the spatial mask accuracy and is a frame size-agnostic metric. It is computed by $\mathcal{J}=\frac{| \hat{\mathbf{P}}_t \cap \mathbf{P}_t |}{| \hat{\mathbf{P}}_t \cup \mathbf{P}_t |}$. Contour accuracy $\mathcal{F}$ estimates whether the contour of predicted mask $\hat{\mathbf{P}}$ is similar with that of ground-truth mask $\mathbf{P}$. From a contour perspective, one can interpret $\hat{\mathbf{P}}$ and $\mathbf{P}$ as a set of closed contours $\mathcal{C}(\hat{\mathbf{P}})$ and $\mathcal{C}(\mathbf{P})$ delimiting the spatial extent of the mask. So one can compute the contour-based precision $P_c$ and recall $R_c$ via a bipartite graph matching. The $\mathcal{F}$ measure is a harmonic value, \ie, $\mathcal{F}=\frac{2P_c R_c}{P_c+R_c}$.  %a bipartite graph matching \cite{martin-pami2004-bgm}.

$\bar{\mathcal{J}}$ denotes $\mathcal{J}~\text{Mean}$ and $\bar{\mathcal{F}}$ denotes $\mathcal{F}~\text{Mean}$, each of which is the average result over all test videos. Meanwhile, we use the mean value of region similarity and contour accuracy as the overall evaluation metric, \ie, $\overline {\mathcal{J}\&\mathcal{F}}$ ($\mathcal{J}\&\mathcal{F}~\text{Mean}$) over all videos. In addition, we use Frame Per Second (FPS) as the metric to evaluate the inference speed.

\subsection{Experimental Setup}
\textbf{Training Phase}.
The Encoder of LSTA is initialized by the DeepLab v3+ \cite{chen-eccv2018-deeplabv3} model pre-trained on MS COCO, while the other modules are randomly initialized using Xavier. In each iteration, we randomly sample a single frame from each of 4 videos (batch size is 4) as query frame, and its all previous frames are grouped into $N=5$ bins, from each of which we randomly select one frame, resulting in $N$ past frames with temporal relations. When going through all available training videos once, it finishes one epoch. For each frame, it is randomly cropped to $465\times 465\times 3$, while random horizontal flipping and scaling are applied. The maximum iteration number is $7.5e4$, and we adopt the SGD (Stochastic Gradient Descent) optimizer with a momentum of 0.9, a weight decay of $1.5e\text{-}4$, and an initial learning rate of $6e\text{-}3$.  Note that our model is trained on DAVIS2017\cite{pont-arxiv2017-davis17} and YouTube-VOS\cite{xu-arxiv2018-ytbvos} with $5e4$ iterations, and is fine-tuned on DAVIS2017\cite{pont-arxiv2017-davis17} with $2.5e4$ iterations to further boost the generalization ability.

\textbf{Inference Phase}.
For unseen videos, according to Algorithm~\ref{alg:inference}, LSTA sequentially takes the current frame as query frame and previous $N$ frames as past frames without any object prior, and outputs the corresponding object masks. Note that there will be insufficient past frames for the foremost $N$ query frames. For such cases, we use the succeeding frames to compensate for the lacking past frames. In addition, we follow AD-Net \cite{yang-iccv2019-anchor} to use instance pruning to filter out some cluttered background by employing instance bounding boxes.
%

% -------------------------  Performance on DAVIS2016 --------------------------
\begin{table}[!h]
	\centering
	\caption{Performance comparisons on DAVIS2016. The methods in top group are semi-supervised and the rest are unsupervised. att: attention mechanism; flow: optical flow feature;  pp: post-processing; crf: conditional random field skill; ip: instance pruning; `-' means the speed is unavailable; $^\dag$ denotes the fps on RTX 2080Ti GPU.
		\label{table:davis2016-sota}}
	\resizebox{0.7\textwidth}{!} 	{  % resize the entire table 
		\begin{tabular}{lcccccccr}
			\toprule[0.75pt]
			Method &Venue & att & flow & pp & $\overline{\mathcal{J}}$  & $\overline{\mathcal{F}}$   & $\overline {\mathcal{J}\&\mathcal{F}}$  & FPS  \\
			\midrule[0.5pt]
			AGUNet\cite{yin-pr2021-agunet}   &PR'21      &\checkmark &  & & 80.7 & 81.0 & 80.9 & 11.1 \\
			%FAVOS\cite{cheng-cvpr2018-favos}  &CVPR'18 & &  & crf  & 82.4   & 79.5 & 81.0 & 0.6  \\
			FEELVOS\cite{voigtlaender-cvpr2019-feelvos} &CVPR'19 & &  &     & 81.1   & 82.2 & 81.7 & 2.0  \\
			MVOS-OL\cite{xiao-pami2020-mvos} &TPAMI'20 & &  &     & 83.3  & 84.1 & 83.7 & 2.3  \\
			DDEAL\cite{yin-tnnls2021-ddeal}      &TNNLS'21  & & & & 85.1 & 85.7 & 85.4  & 25.0 \\
			\midrule[0.5pt]
			PDB\cite{song-eccv2018-pdb}  &ECCV'18   & &  & crf   & 77.2   & 74.5 & 75.9 & -    \\
			%LSMO\cite{tokmakov-ijcv2019-lsmo} &IJCV'19 & & \checkmark    & crf   & 78.2   & 75.9 & 77.1 & -    \\
			%UOVOS\cite{zhuo-tip2020-uovos} &TIP'20  & & \checkmark & crf & 77.8 & 72.0 & 74.9 & 0.1 \\
			AGNN\cite{wang-iccv2019-agnn} &ICCV'19 & \checkmark &   & crf   & 80.7   & 79.1 & 79.9 & 0.3  \\
			COSNet\cite{lu-cvpr2019-cosnet} &CVPR'19 & \checkmark &  & crf   & 80.5   & 79.5 & 80.0   & 1.2    \\
			STEm-Seg\cite{athar-eccv2020-stem} &ECCV'20  & & &  & 80.6   & 80.6 & 80.6 & 0.7  \\
			AD-Net\cite{yang-iccv2019-anchor} &ICCV'19 & \checkmark &   & ip    & 81.7   & 80.5 & 81.1 & 0.3  \\
			MATNet\cite{zhou-tip2020-matnet} &TIP'20 & \checkmark & \checkmark & crf   & 82.4   & 80.7 & 81.6 & 1.3  \\
			FSNet\cite{ji-iccv2021-fsnet} &ICCV'21 & \checkmark & \checkmark   &     & 82.3   & 83.3 & 82.8 & 12.5 \\
			$\text{FSNet}^{\ast}$\cite{ji-iccv2021-fsnet} &ICCV'21 & \checkmark & \checkmark  & crf   & 83.4   & 83.1 & 83.3 & -    \\
			3DC\cite{mahadevan-bmvc2020-3dc}  &BMVC'20 & &   &     & 83.9   & 84.2  & 84.1 & 4.5  \\
			DASPP\cite{zhao-pr2021-realtime} &PR'21 & &  &     & 63.4   & 60.2 &61.8  & 29.4  \\
			%IKE\cite{haller-tpami2021-iterative} &TPAMI'21 & &\checkmark &     & 70.8   & 68.6 &  69.7 & -  \\
			AGS\cite{wang-pami2021-ags} &TPAMI'21 & \checkmark &     & crf   & 79.7   & 77.4 & 78.6 & -    \\
			RTNet\cite{ren-cvpr2021-rt} & CVPR'21  & \checkmark & \checkmark   &    & 85.6 & 84.7  & 85.2 & 4.3 \\
			OFS\cite{meunier-tpami2022-ofs} & TPAMI'23 & &\checkmark   &     & 69.3   & 70.7 &70.0  & -  \\
			FEM-Net\cite{zhou-tcsvt2021-fem}  & TCSVT'22   &\checkmark & \checkmark & & 79.9 & 76.9 & 78.4 &  16.0 \\
			IMCNet\cite{xi-tcsvt2022-imcn} & TCSVT'22 &\checkmark &  &     & 82.7   & 81.1 &81.9  & -  \\
			IMP\cite{lee-aaai2022-imp} & AAAI'22  &\checkmark   &   &  & \underline{84.5}  & \textbf{86.7} & \textbf{85.6} & 	1.79$^\dag$  \\
			TMO\cite{cho-wacv2023-tmo} & WACV'23 &   & \checkmark  &	  & \textbf{85.6} &\underline{86.6}  &\textbf{86.1}  & 24.8  \\
			\midrule[0.5pt]
			LSTA (Ours)  &  & \checkmark &   &     & 82.4   & 84.3 & 83.4 & \textbf{42.8} \\
			$\text{LSTA}^{*}$(Ours) & & \checkmark &   &  ip   & 82.7   & 84.8 & 83.8 &  \underline{36.2} \\
			\toprule[0.75pt]
		\end{tabular}
	}
\end{table}

\subsection{Quantitative Results}
We show the quantitative comparison results on DAVIS2016, DAVIS2017, YouTube-Objects, and FBMS, with rigorous analysis in the following.

% --------------------  Computational analysis on DAVIS2016 ---------
\begin{table}[!t]
	\centering
  \caption{Computational analysis on DAVIS2016.}
		\label{tbl:speed_modelPara}  % caption and label should be paired 
	%\resizebox{\textwidth}{!} 	{  % resize the entire table 
	\setlength{\tabcolsep}{0.6mm}{  % control the table size on a whole
		\begin{tabular}{l c c c c  c  c c c c}
			\toprule[0.75pt]
			\multirow{2}{*}{Method} & \multirow{2}{*}{Venue} & \multirow{2}{*}{Backbone} & \multirow{2}{*}{$\overline {\mathcal{J}\&\mathcal{F}}$} &\multirow{2}{*}{\scriptsize{Params(M)$\downarrow$}} & \multirow{2}{*}{\scriptsize{FLOPs(G)$\downarrow$}}  & \multicolumn{2}{c}{Speed (FPS)} \\
			\cmidrule{7-8}
			&  &  &  &  &  & TITAN Xp$\uparrow$  & 2080Ti $\uparrow$ \\
			\midrule[0.5pt]  		
			 MATNet\cite{zhou-tip2020-matnet}  & TIP'20  & ResNet101 & 81.6 & 142.7 & \textbf{193.7} & 1.3 & 1.8 \\
			COSNet\cite{lu-cvpr2019-cosnet} & CVPR'19 & DeepLabv3 & 80.0 & 81.2 & 585.5 & 1.2 & 1.6 \\
			RTNet\cite{ren-cvpr2021-rt}     & CVPR'21 &ResNet101 & \underline{85.2}  & 277.2 &489.6 & 4.3 & 5.8 \\
			IMCNet\cite{xi-tcsvt2022-imcn}  & TCSVT'22 &ResNet101 & 81.9 & \textbf{47.9} & 391.8 & 11.0 & \underline{19.0} \\
			TMO\cite{cho-wacv2023-tmo}      & WACV'23 &ResNet101 & \textbf{86.1} & 93.0  & \underline{344.9} & \underline{24.8}  & \underline{39.1} \\
			\midrule[0.5pt]
			LSTA(Ours)   & -     & DeepLabv3 &  83.4 & \underline{60.5}  &349.5  &  \textbf{42.8}  & \textbf{62.3}  \\
			\toprule[0.75pt]
		\end{tabular}
	}
\end{table}

% ------ Component analysis of LSTA ----------
\begin{table}[!t]
	\centering
	\caption{Component computational analysis on DAVIS2016.}
		\label{tbl:component_complexity}	
	\setlength{\tabcolsep}{1.4mm}{
		\begin{tabular}{cccccc}
			\toprule[0.75pt]
			Blocks         & Encoder  &  LTM    &  STA   & Decoder  & LSTA        \\ 
			\midrule[0.5pt]  
			FLOPs(G)       & 279.5    & 4.2     & 2.4    & 63.4     &349.5   \\
			Params(M)      & 59.2     & 0.03    & 0.02   & 1.2      &60.5    \\
			\toprule[0.75pt]
		\end{tabular}
	}	
\end{table}

\textbf{DAVIS2016}. The results of our LSTA model and a number of state-of-the-art alternatives are recorded in Table~\ref{table:davis2016-sota}. Among them, the above seven methods are semi-supervised models, including FEELVOS (Fast End-to-End Embedding Learning) \cite{voigtlaender-cvpr2019-feelvos}, AGUNet (Annotation-Guided U-Net)\cite{yin-pr2021-agunet}, MVOS-OL (Meta VOS Online Learning)\cite{xiao-pami2020-mvos}, and DDEAL (Directional Deep Embedding and Appearance Learning) \cite{yin-tnnls2021-ddeal}.  The remaining ones are all unsupervised models, including PDB (Pyramid Dilated Bidirectional ConvLSTM)\cite{song-eccv2018-pdb}, AGS (Attention-Guided object Segmentation)\cite{wang-pami2021-ags}, AGNN (Attentive Graph Neural Networks)\cite{wang-iccv2019-agnn}, COSNet (Co-Attention Siamese Networks)\cite{lu-cvpr2019-cosnet}, STEm-Seg (Spatio-Temporal Embeddings for instance Segmentation) \cite{athar-eccv2020-stem}, AD-Net (Anchor Diffusion Network)\cite{yang-iccv2019-anchor}, MATNet (Motion-Attentive Transition Network)\cite{zhou-tip2020-matnet}, FSNet (Full-duplex Strategy Network)\cite{ji-iccv2021-fsnet}, FEM-Net (Flow Edge-based Motion-attentive Network)\cite{zhou-tcsvt2021-fem}, 3DC (3D Convolutions)\cite{mahadevan-bmvc2020-3dc}, RTNet (Reciprocal Transformation Network)\cite{ren-cvpr2021-rt}, IMCNet (Implicit Motion-Compensated Network) \cite{xi-tcsvt2022-imcn}, OFS (Optical Flow Segmentation), \cite{meunier-tpami2022-ofs}, DASPP (Dynamic Astrous Spatial Pyramid Pooling) \cite{zhao-pr2021-realtime}, IMP (Iterative Mask Propagation)\cite{lee-aaai2022-imp}, and TMO (Treating Motion as Option)\cite{cho-wacv2023-tmo}. Among them, many methods such as OFS, FEM-Net, RT-Net and TMO, adopt the two-stream VOS framework that employs optical flow as the motion modality to capture the temporal relations; IMCNet aligns motion information from nearby frames to the current frame, and adopts a co-attention mechanism to learn the global representation; IMP repeats easy frame selection with mask propagation but the inference speed is very low.

% semi-supervised method: FAVOS (Online) \cite{cheng-cvpr2018-favos}
% unsupervised method: LSMO (Learning to Segment Moving Objects) \cite{tokmakov-ijcv2019-lsmo}, IKE (Iterative Knowledge Exchange) \cite{haller-tpami2021-iterative}, UOVOS (Unsupervised Online)\cite{zhuo-tip2020-uovos}, 

These records show that our LSTA model enjoys more satisfying overall performance, and very competitive with SOTA unsupervised alternatives in terms of $\mathcal{F}$ Mean, \ie, $84.3\%$. Especially, LSTA ranks Top-1 in the inference speed, achieving 42.8 fps, almost 9.5 times of that of the best candidate 3DC. This demonstrates that LSTA can be deployed in highly demanding applications with its real-time segmentation ability. Besides, we have the following observations:
\begin{itemize}
	\item Our LSTA can beat against several semi-supervised VOS methods, such as FAVOS and RGMP, in terms of both region similarity and contour accuracy. Meanwhile, LSTA is much faster ($1.7\times$) than DDEAL who has the best performance in semi-supervised setting.
	
	\item Most of the unsupervised VOS methods adopt post-processing techniques, such as conditional random field or instance pruning to refine the object mask. We also use instance pruning to slightly upgrade the performance by $0.4\%$ of $\mathcal{J}\&\mathcal{F}~\text{Mean}$.
	
	\item Unlike existing unsupervised VOS methods that use costly optical flow features, our model attempts to discover the patterns of constantly present and moving objects, by encoding both global and local spatiotemporal correlations across frames with the long-short temporal attention mechanism.
	
	\item While the latest work TMO achieves the best segmentation performance, it requires the expensive optical flow features and the inference speed is much lower than ours, \ie, 24.8fps versus 42.8fps. This demonstrates the LSTA strikes a good balance between the performance and speed.
	
\end{itemize}

In addition, we provide the computational comparison and component computational analysis in Table~\ref{tbl:speed_modelPara} and Table~\ref{tbl:component_complexity}, respectively. Note that we also provide the inference speed using 2080Ti card for reference. It can be seen from the tables that our LSTA achieves the best inference speed at 42.8fps using TITAN Xp, which is almost the twice faster than that of the second best one. Meanwhile, the developed LTM module and STA module are very lightweight in terms of FLOPs with the negligible parameter size.

% -------------------------  Performance on DAVIS2017 --------------------------
\begin{table}[!h]
	\centering
	\caption{Performance comparisons on DAVIS2017.}
		\label{table:davis2017}
	\begin{tabular}{lccccccc}
		\toprule[0.75pt]
		Method  &Venue  & att & flow & pp & $\overline{\mathcal{J}}$ &  $\overline{\mathcal{F}}$  & $\overline {\mathcal{J}\&\mathcal{F}}$ \\
		\midrule[0.5pt]
		RVOS\cite{ventura-cvpr2019-rvos} &CVPR'19  & &  &     & 36.8                & 45.7                & 41.2                    \\
		PDB\cite{song-eccv2018-pdb} &ECCV'18  & &  &  crf   & 53.2                & 57.0                  & 55.1                    \\
		MATNet\cite{zhou-tip2020-matnet} &TIP'20 & \checkmark & \checkmark  &  crf   & 56.7                & 60.4                & 58.6                    \\
		STEm-Seg\cite{athar-eccv2020-stem} &ECCV'20 & & &     & 61.5                & 67.8                & 64.7                    \\
		UnOVOST\cite{luiten-wacv2020-unovost} &WACV'20  & & \checkmark &     & 66.4    & 69.3     & 67.9    \\
		AGS\cite{wang-pami2021-ags}  & TPAMI'21  &\checkmark &  &   crf  & 55.5                & 59.5      & 57.5          \\
		DyStaB\cite{yang-cvpr2021-dystab} &  CVPR'21  &  & \checkmark & crf & 58.9  & - & 58.9 \\
		TAODA\cite{zhou-cvpr2021-taoda}  &  CVPR'21  & & & & 63.7  & 66.2 & 65.0 \\
		\midrule[0.5pt]
		LSTA (Ours)     & &\checkmark & &    & \textbf{70.8}                & \textbf{75.8}                & \textbf{73.3}                    \\
		LSTA$^\dagger$ (Ours)  &  &\checkmark & &    & \underline{67.8}                & \underline{72.3}                & \underline{70.1}                    \\
		\toprule[0.75pt]
	\end{tabular}
\end{table}

\textbf{DAVIS2017}. This dataset is much more difficult for segmentation, since there may exist multiple objects in a single video. To handle this case, some previous unsupervised VOS works, such as UnOVOST (Unsupervised Offline VOS) \cite{luiten-wacv2020-unovost}, MATNet (Motion-Attentive Transition Network) \cite{zhou-tip2020-matnet}, AGS (Attention-Guided object Segmentation) \cite{wang-pami2021-ags}, employ instance segmentation model Mask R-CNN \cite{he-pami2020-maskrcnn} to obtain object proposals involving mask and boundary box. Unlike them, TAODA (Target-Aware Object Discovery and Association) \cite{zhou-cvpr2021-taoda} introduces an instance discrimination network to obtain object proposals in a bottom-up fashion. Moreover, DyStaB (Dynamic-Static Bootstrapping) \cite{yang-cvpr2021-dystab} employs a motion segmentation module to perform temporally consistent region separation, and it requires the expensive optical flow features and CRF post-processing to get final results. Note that STEm-Seg \cite{athar-eccv2020-stem} used spatiotemporal embeddings to find instances, which is based on Gaussian distribution, failing to handle complex appearance of objects. The earlier works, PDB (Pyramid Dilated Bidirectional ConvLSTM) \cite{song-eccv2018-pdb} and RVOS (Recurrent network) \cite{ventura-cvpr2019-rvos} identify the instance by recurrent neural networks with ConvLSTM, which fails to encode long-range temporal relations. Similarly, we use HTC (Hybrid Task Cascade) \cite{chen-cvpr2019-htc} to obtain object proposals of the first frame, which are then processed to yield object mask, and adopt STCN (Space-Time Correspondence Network) \cite{cheng-nips2021-stcn} to obtain initial object masks of subsequent frames in a semi-supervised manner. After that, we use our LSTA model to obtain predicted masks that contain multiple objects. In addition, we follow UnOVOST using Mask R-CNN to extract object proposals and its merging strategy with the pixel probability obtained by our model, and the results are listed in the bottom row.

The comparison results of the above methods are tabulated in Table~\ref{table:davis2017}, which has shown the significant improvements brought by our model with good generalization ability. For example, LSTA achieves $70.8\%$ on $\mathcal{J}~\text{Mean}$, $75.8\%$ on  $\mathcal{F}~\text{Mean}$, and $73.3\%$ on $\mathcal{J}\&\mathcal{F}~\text{Mean}$, which have improvements of $4.4\%$, $6.5\%$, and $5.4\%$ compared to the most competitive alternative, \ie, UnOVOST. We attribute this to the fact that our approach is able to encode both long-range and shot-range spatiotemporal pixel-wise relations of the current frame and the past frames, which helps to better capture constantly present objects and moving objects in video.

\begin{table*}[!h]
	\centering
	\caption{Performance comparisons on YouTube-Objects with 10 categories. The number of videos in each category is in parenthesis.}
		\label{table:youtube-objects}
	\resizebox{\textwidth}{!} 	{  % resize the entire table 
		\setlength{\tabcolsep}{0.8mm}{  % control the table size on a whole
			\begin{tabular}{lccccccccccc}
				\toprule[0.75pt]
				Method & SegFlow\cite{cheng-iccv2017-sfl} & PDB\cite{song-eccv2018-pdb}  & MATNet\cite{zhou-tip2020-matnet} & COSNet\cite{lu-cvpr2019-cosnet} & AGNN\cite{wang-iccv2019-agnn} & AGS\cite{wang-pami2021-ags} & RTNet\cite{ren-cvpr2021-rt} &IMCNet\cite{xi-tcsvt2022-imcn} 
				& TMO\cite{cho-wacv2023-tmo}   & LSTA                          \\
				Venue &ICCV'17 &ECCV'18  &TIP'20  &CVPR'19 &ICCV'19 & TPAMI'21 &  CVPR'21 & TCSVT'22 &  WACV'23 & Ours  \\
				\midrule[0.5pt]
				att &  & & \checkmark & \checkmark & \checkmark &\checkmark & \checkmark &\checkmark & & \checkmark \\
				flow &\checkmark & 	&\checkmark & & & &\checkmark& &\checkmark  & \\
				pp & & crf &  crf & crf & crf & crf & & & &  \\
				\midrule[0.5pt]
				Airplane(6)        & 65.6   & 78.0      & 72.9    & 81.1    & 81.1 & \textbf{87.7}  & 84.1 & 81.1 & \underline{85.7} & 85.1    \\
				Bird(6)              & 65.4   & 80.0           & 77.5    & 75.7  & 75.9 & 76.7 & \underline{80.2}  &\textbf{81.1} & 80.0  & 75.9     \\
				Boat(15)           & 59.9   & 58.9          & 66.9    & \underline{71.3}   & 70.7 & \textbf{72.2}  & 70.1  &70.3 & 70.1 & 66.5    \\
				Car(7)               & 64.0   & 76.5      & \underline{79.0}    & 77.6    & 78.1  & 78.6 & \textbf{79.5}  &77.1  & 78.0 & 78.4    \\
				Cat(16)             & 58.9   & 63.0         & \textbf{73.7}     & 66.5      & 67.9 & 69.2  & 71.8 &73.3 & \underline{73.6} & 72.4 \\
				Cow(20)            & 51.2   & 64.1   & 67.4    & 69.8    & 69.7 & 64.6  & \underline{70.1} &66.8 & \textbf{70.3} & 67.1  \\
				Dog(27)            & 54.1   & 70.1      & 75.9     &  76.8    & \underline{77.4} & 73.3    & 71.3 &74.8 & 76.8 & \textbf{77.9} \\
				Horse(14)         & 64.8    & \underline{67.6}         & 63.2      &  67.4   & 67.3 & 64.4  & 65.1 &64.8 & 66.2 & \textbf{68.5} \\
				Motorbike(10)   & 52.6    & 58.4       & 62.6    & \underline{67.7}  & \textbf{68.3} & 62.1 & 64.6 &58.7 & 58.6  & 65.5     \\
				Train(5)           & 34.0   & 35.3      & 51.0     & 46.8   & 47.8 & 48.2   & 53.3 &\underline{56.8} & 47.0 & \textbf{57.5}   \\
				\midrule[0.5pt]
				$\overline{\mathcal{J}}$   & 57.1   & 65.5  & 69.0   & 70.5   & 70.8  & 69.7  & \underline{71.0} &70.5 &  \textbf{71.5} & \textbf{71.5}    \\
				FPS   & -  & -   & -     & 0.9    & 0.4 &  4.2    & - &- &  \underline{18.5} & \textbf{35.5}    \\
				\toprule[0.75pt]
			\end{tabular}
		}
	}
\end{table*}

\textbf{YouTube-Objects}. Following previous work\cite{zhou-tip2020-matnet}, we give segmentation results in terms of $\mathcal{J}~\text{Mean}$ for each of the \emph{ten} semantic categories, as in Table~\ref{table:youtube-objects}. The compared methods include SFL\cite{cheng-iccv2017-sfl}, PDB\cite{song-eccv2018-pdb}, MATNet\cite{zhou-tip2020-matnet}, AGS\cite{wang-pami2021-ags}, COSNet\cite{lu-cvpr2019-cosnet}, AGNN\cite{wang-iccv2019-agnn}, RTNet\cite{ren-cvpr2021-rt}, IMCNet \cite{xi-tcsvt2022-imcn}, and TMO\cite{cho-wacv2023-tmo}. Among them, our model outperforms the rest ones in overall performance, achieving $71.5\%$ on $\mathcal{J}~\text{Mean}$ at $35.5$ fps in inference speed. Especially, the performance of our method is comparable to the latest TMO model which employs the more complicated framework, and our speed is twice faster. The encouraging records have verified the advantage of our approach in terms of both effectiveness and efficiency. 
	
% -------------------------  Performance on FBMS --------------------------
\begin{table}[!h]
	\centering
	\caption{Comparison results on FBMS validation set.}
		\label{table:fbms}
	\resizebox{\textwidth}{!}{
		%\small
		%\scalebox{0.8}{
		\setlength{\tabcolsep}{0.8mm}{
			\begin{tabular}{lcccccccc}
				\toprule[0.75pt]
				Method  & OBN\cite{li-eccv2018-obn}  & PDB\cite{song-eccv2018-pdb} & COSNet\cite{lu-cvpr2019-cosnet} & MATNet\cite{zhou-tip2020-matnet}   &OFS\cite{meunier-tpami2022-ofs} & AGS\cite{wang-pami2021-ags} &DASPP\cite{zhao-pr2021-realtime}    & LSTA                          \\
				Venue  &ECCV'18 &ECCV'18 &CVPR'19 &TIP'20 & TPAMI'23  & TPAMI'21 & PR'21& Ours \\
				\midrule[0.5pt]
				att  & & & \checkmark & \checkmark & &\checkmark  &\checkmark &\checkmark \\
				flow   &\checkmark & & &\checkmark &\checkmark & & & \\
				pp   &crf & crf & crf &crf & &crf & &  \\
				\midrule[0.5pt]
				$\overline{\mathcal{J}}$      & 73.9 & 74  & 75.6   & \underline{76.1} &57.8 & 76.0 &62.3 & \textbf{77.3} \\
				FPS                           &  \underline{4.7}  & -   & 0.9 & 0.4  &- &- &- & \textbf{41.4} \\
				\toprule[0.75pt]
			\end{tabular}
		}
	}
\end{table}

% IET (Instance Embedding Transfer)\cite{li-cvpr2018-iet}, 
\textbf{FBMS}. We compare our model with OBN (Object Bilateral Networks)\cite{li-eccv2018-obn}, PDB\cite{song-eccv2018-pdb}, COSNet\cite{lu-cvpr2019-cosnet}, MATNet\cite{zhou-tip2020-matnet}, OFS \cite{meunier-tpami2022-ofs}, AGS\cite{wang-pami2021-ags}, and DASPP \cite{zhao-pr2021-realtime} methods on FBMS, whose results are shown in Table~\ref{table:fbms}. From these records, we see our LSTA method gets the highest region similarity, \ie, $77.3\%$, with a margin of $1.2\%$ compared to the second best, \ie, MATNet. Meanwhile, our method achieves a satisfying balance between segmentation performance and inference speed without optical flow features and post-processing, \eg, its inference speed is 41.4fps, which helps to being deployed in highly-demanding environment.

\subsection{Ablation Study}
This section makes extensive analysis on the contribution of individual components of LSTA, the number of past frames $N$ and the strategy of selecting them, the patch size and stride in STA block, the influences of channel numbers $c$ and whether sharing Conv2D layers, the tradeoff parameter $\alpha$, and the performance on the videos with various visual characteristics on DAVIS2016.

% -------------------------  ablation study: STA-LTM-L2 --------------------------
\begin{table}[!t]
	\centering
	\caption{Ablation study of individual components on DAVIS2016. }
	\label{table:ablation-blocks}
	\begin{tabular}{lcccc}
		\toprule[0.75pt]
		Blocks   & $\mathcal{J}$ Mean                     & $\mathcal{F}$ Mean                       &  $\overline{\mathcal{J\& F}}$       & Gain            \\
		\midrule[0.5pt]	
		Baseline	    & 76.3	     &79.6             &77.9      &  -  \\
		+LTM         & 80.1        & 82.3            & 81.2     & +3.3                          \\
		+STA         & 79.5        & 82.5            & 81.0     & +3.1                   \\
		+LTM+STA  & 81.4        & 83.8           & 82.6      & +4.7                   \\
		+LTM+STA+$\mathcal{L}_2$    & \textbf{82.4}  & \textbf{84.2} & \textbf{83.3} & +5.4 \\ 	
		\toprule[0.75pt]
	\end{tabular}
\end{table}

%----------------------------------- Ablation visualization results: DAVIS2016  ---------------------------------------
\begin{figure}[!t]
	\centering
	\includegraphics[width=0.7\textwidth]{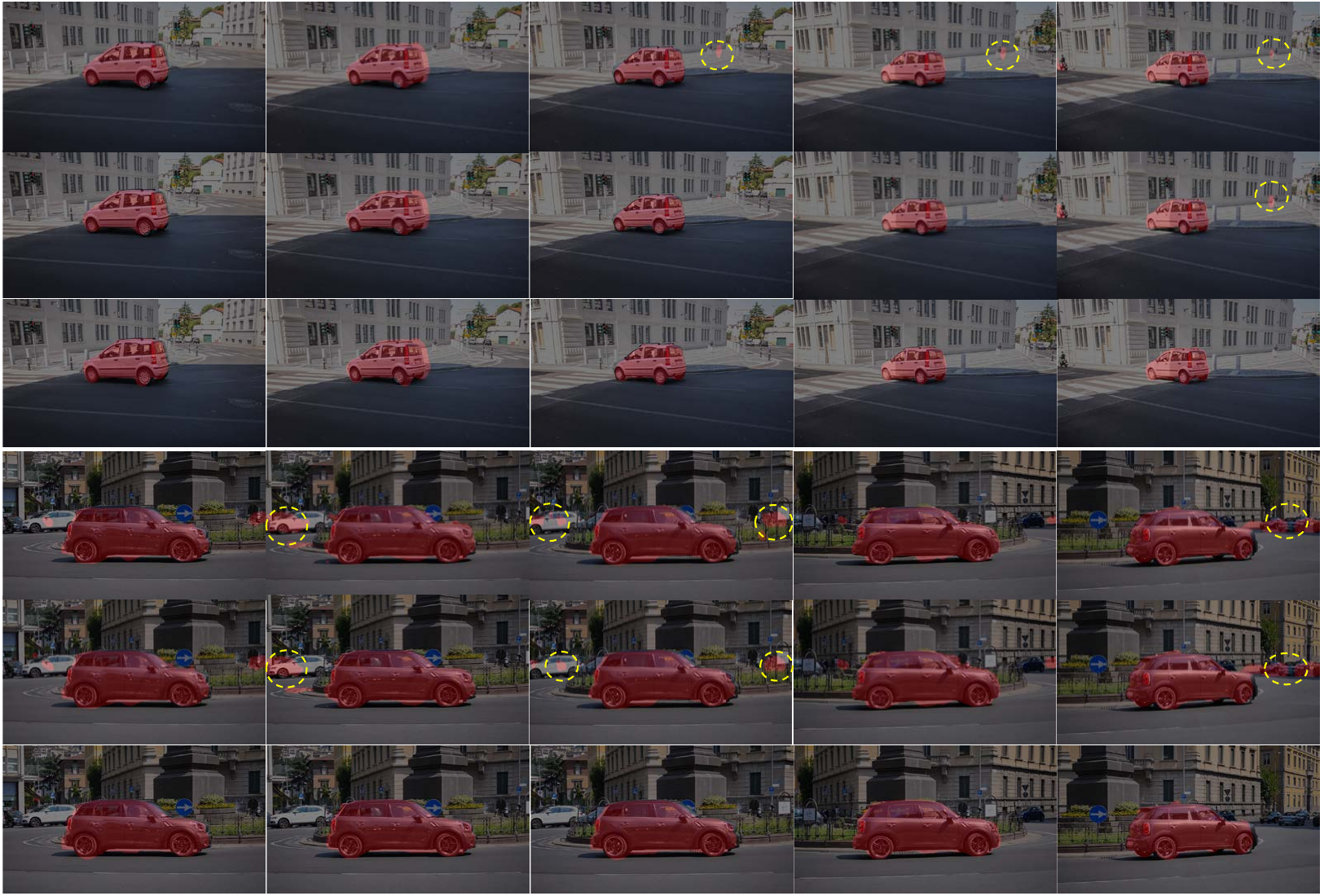}
	\caption{Visualization results of the ablation study on individual components. Row~1/3: LTM, Row~2/4: LTM+STA, Row~3/6: LTM+STA+$\mathcal{L}_2$. The dashed yellow circle highlights the difference.
	\label{fig:ablation_vis}}
\end{figure}

\textbf{Individual components}. We show the effects of different components in Table~\ref{table:ablation-blocks}, which examines the Light Temporal Attention block and the Short Temporal Attention block as well as $\mathcal{L}_2$ loss terms. For baseline, we use $\mathcal{L}_1$ loss term. From the table, we observe that both LTM and STA promote the performance by $3.8\%$ (row~2) and $3.2\%$ (row~3), respectively, on $\mathcal{J}~\text{Mean}$, compared to baseline. Besides, the coupling of LTM and STA brings about an upgrade (row~4) by some margin of $0.6\%$ and $3.2\%$ on $\mathcal{J}\&\mathcal{F}~\text{Mean}$, respectively. This demonstrates the necessity of simultaneously considering both global and local spatiotemporal pixel relations of the current frame and the past frames. In addition, the knowledge distillation skill is beneficial for boosting the performance by $1.0\%$ (bottom row) on $\mathcal{J}~\text{Mean}$.  Meanwhile, we show some visualization examples of using different components in Fig.~\ref{fig:ablation_vis}. As shown in the figure, the error area becomes smaller or disappears when using both STA module and LTM module (Row~2 and 4); meanwhile, the segmentation quality is further improved by adding the knowledge distillation loss $\mathcal{L}_2$ (Row~3 and 6).

%----------------------------  ablation study: LTM past frames number and selection  ------------------------------------------
\begin{table}[!t]
	\begin{minipage}{0.48\linewidth}
		\centering
		\caption{Number $N$ of past frames.}
		\label{table:num_past_frams}
		\scalebox{0.9}{
			\begin{tabular}{cccc}
				\toprule[0.75pt]
				$N$ & $\overline{\mathcal{J}}$ & $\overline{\mathcal{F}}$  & $\overline {\mathcal{J}\&\mathcal{F}}$ \\
				\midrule[0.5pt]
				2              & 82.0               & 84.1               & 83.1                       \\
				3              & 82.1               & 84.1               & 83.1                       \\
				5              & \textbf{82.4}               & \textbf{84.2}               & \textbf{83.3}                       \\
				7              & 82.1               & 84.1               & 83.1                       \\
				9              & 81.9               & 84.1               & 83.0                       \\
				11            & 81.9               & 84.0               & 83.0                       \\ 		
				\toprule[0.75pt]
			\end{tabular}
		}
	\end{minipage} % no newline
	\begin{minipage}{0.48\linewidth}
		\centering
		\caption{Past frame selection ($N$=5).}
		\label{table:select_past_frames}
		\scalebox{0.9}{
			\begin{tabular}{lccc}
				\toprule[0.75pt]
				Frames   & $\overline{\mathcal{J}}$             & $\overline{\mathcal{F}}$              & $\overline{\mathcal{J}\&\mathcal{F}}$               \\
				\midrule[0.5pt]
				first             & 81.2                           & 83.5                           & 82.4                           \\
				prev          & 82.0                           & 84.1                           & 83.0                           \\
				first\&prev & 81.4                           & 83.8                           & 82.6                           \\
				rand $N$          & 81.9                           & 84.0                           & 82.9                           \\
				every $N$           & 81.9                           & 84.1                           & 83.0                           \\
				prev $N$        & \textbf{82.4} & \textbf{84.3} & \textbf{83.4} \\ 	\toprule[0.75pt]
			\end{tabular}
		}
	\end{minipage}
\end{table}

\textbf{Number of past frames in LTM}. We vary the number of past frames from 2 to 11 for LTM, and the results are shown in Table~\ref{table:num_past_frams}. It can be seen that when $N$ is 5, the model achieves the best performance, which suggests that much more past frames can not provide additional spatiotemporal information due to frame redundancy.

\textbf{Selecting past frames}. Table~\ref{table:select_past_frames} gives the results of six ways of selecting the past frames. Selecting the nearest frame (row~2) performs better than using the first frame (row~1). Besides, using both the first frame and the previous one will slightly degrade performance in comparison of using only the previous one, which might be reason that the object mask changes a lot with time and the first frame may mislead mask prediction. When using more past frames, \eg, two frames in row~3 and five frames in the last three rows, the segmentation performance is improved, especially using previous $N$ nearest frames (bottom row) in inference phase.

%--------------------------- ablation study: patch size and stride in STA ----------------------------------------

\begin{table}[!t]
	\centering
	\caption{Patch size $k$ and stride $d$ in STA block.}
	\label{table:tla_batchsize_stride}
	\begin{tabular}{ccccc}
		\toprule[0.75pt]
		$k$ & $d$ & $\mathcal{J}$~Mean            & $\mathcal{F}$~Mean           & $\mathcal{J}\&\mathcal{F}$~Mean         \\
		\midrule[0.5pt]
		2 & 1               & 82.0                           & \textbf{84.2} & 83.1                           \\
		4 & 2               & 82.0                           & \textbf{84.2} & 83.1                           \\
		8 & 4               & \textbf{82.4} & \textbf{84.3} & \textbf{83.4} \\
		16&8              & 82.0                           & 84.1                           & 83.0                           \\
		32&16             & 81.9                           & 83.9                           & 82.9                           \\ 		
		\toprule[0.75pt]
	\end{tabular}
\end{table}

\textbf{Patch size and stride in STA}. STA adopts the locality-based sliding window strategy to reduce the computational cost, which is largely governed by the patch size $k$ and the stride $d$. We vary them from 2 to 32 and 1 to 16, respectively, and the results are recorded in Table~\ref{table:tla_batchsize_stride}. From the table, STA block works the best when the batch size is 8 and the stride is 4, which are both modest values. Larger patches or smaller ones do not help the improvements of capturing local spatiotemporal pixel relations of the current frame and the previous frame.

%----------------------------------- ablation study: number of channels and sharing Conv2D ---------------------------------------
\begin{table}[!t]
	\begin{minipage}{0.48\linewidth}
		\centering
		\caption{Channel Number $c$.}
		\label{table:num_channel}
		\scalebox{1}{
			\begin{tabular}{lccc}
				\toprule[0.75pt]
				$c$ & $\overline{\mathcal{J}}$            &  $\overline{\mathcal{F}}$             & $\overline {\mathcal{J}\&\mathcal{F}}$ \\
				\midrule[0.5pt]
				16             & 79.4                          & 80.9                           & 80.2                             \\
				32             & 81.4                           & 83.7                           & 82.6                             \\
				64             & \textbf{82.4}       & \textbf{84.3}        & \textbf{83.4}   \\
				128            & 81.8                           & 84.0                             & 82.9                             \\
				256            & 81.5                           & 83.9                           & 82.7                             \\
				\toprule[0.75pt]
			\end{tabular}
		}
	\end{minipage}  % without newline
	\begin{minipage}{0.48\linewidth}
		\centering
		\caption{Sharing Conv2D layer.}
		\label{table:sharingConv2D}
		\scalebox{1}{
			\begin{tabular}{lccc}
				\toprule[0.75pt]
				Share & $\overline{\mathcal{J}}$ & $\overline{\mathcal{F}}$  & $\overline {\mathcal{J}\&\mathcal{F}}$ \\
				\midrule[0.5pt]
				$\psi=\phi$              & 81.5               & 83.9               & 82.7                       \\
				$\psi=\theta$            & 81.6               & 84.2               & 82.9                       \\
				$\phi=\theta$            & 81.7               & 84.1               & 82.9                       \\
				$\psi=\phi=\theta$       & 81.6               & 84.1               & 82.9                       \\
				None                       & \textbf{82.4}               & \textbf{84.3}               & \textbf{83.4}                       \\ 		
				\toprule[0.75pt]
			\end{tabular}
		}
	\end{minipage}
\end{table}

\textbf{Number of channels in Conv2D}. Table~\ref{table:num_channel} shows the results of increasing the number of channels ($c$) in Conv2D from 16 to 256. The region similarity metric is improved by $3.2\%$ using 64 channels, compared to that with 16 channels. More channels encode more accurate spatial structure of frame data, but the performance degrades when $c$ is over 100. This might because some noise in channel dimension is mixed with the feature maps.

\textbf{Sharing Conv2D layer}. Our model adopts Conv2D layers for encoding the past frames in LTM by $\phi(\cdot)$, the previous frame in STA by $\theta(\cdot)$, and the current frame in STA by $\psi(\cdot)$. So we explore the influences of sharing them in various forms as in Table~\ref{table:sharingConv2D}. When sharing either two of them or three all, the performance is worse than using Conv2D layers independently for them. It suggests that learning parameters independently for them can model spatial pixel correlations better.

%----------------------------------- ablation study: various appearance features ---------------------------------------
\begin{figure}[!t]
	\centering
	\includegraphics[width=0.7\textwidth]{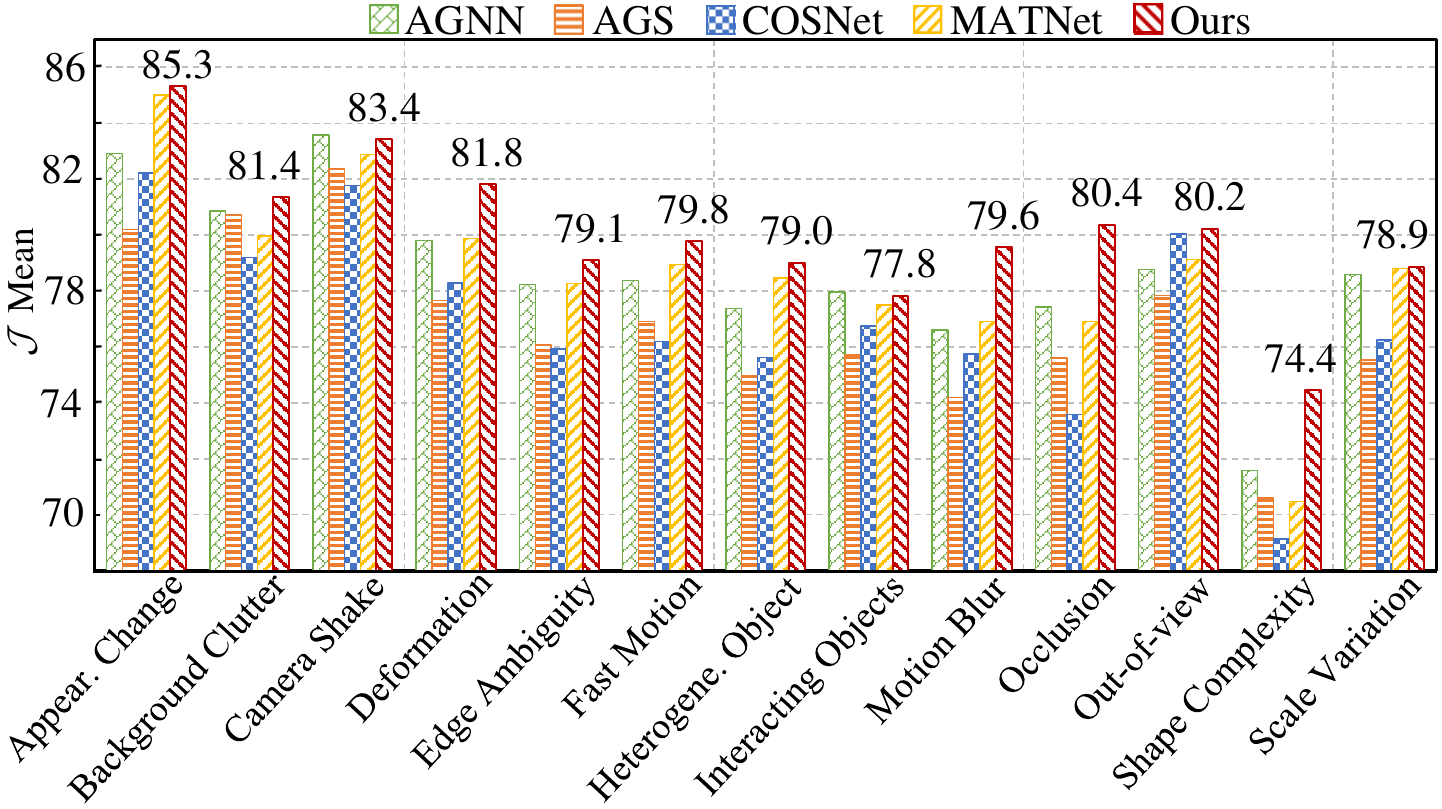}
	\caption{Performance on videos with various visual characteristics.}
	\label{fig:attribute}
\end{figure}

\textbf{Tradeoff parameter $\alpha$}. We vary the tradeoff parameter $\alpha$ in the loss function from 0.1 to 0.9 at an interval of 0.2, and the results are shown in Table~\ref{tbl:alpha}. The results show that the performance of our method tends to rise up before 0.5 and saturates after the best value 0.5. This indicates that our method performs best when the model loss $\mathcal{L}_1$ and the knowledge distillation loss $\mathcal{L}_2$ contribute equally to the objective function. 
	
% ----------------  Ablations on DAVIS2016 ----------------
\begin{table}[!h]
	\centering
	\caption{Tradeoff parameter $\alpha$ in the loss.}
	\label{tbl:alpha}
	\setlength{\tabcolsep}{1.4mm}{
		\begin{tabular}{ccccccccccccccc}
			\toprule[0.75pt]
			$\alpha$ & $\overline{\mathcal{J}}$ &  $\overline{\mathcal{F}}$  & $\overline {\mathcal{J}\&\mathcal{F}}$  \\ 
			\midrule[0.5pt]  
			0.1      & 78.3 & 80.2  & 79.3    \\
			0.3      & 80.1 & 81.5  & 80.8    \\
			0.5      & \textbf{82.4} & \textbf{84.3}  & \textbf{83.4}    \\
			0.7      & 81.9 & 82.8  & 82.4    \\
			0.9      & 81.8 & 83.0  & 82.4    \\
			\toprule[0.75pt]
		\end{tabular}
	}	
\end{table}
	
\textbf{Various visual characteristics}. We show the LSTA performance on video data with 13 kinds of visual characteristics in Fig.~\ref{fig:attribute}. As depicted in this histogram, our model consistently performs better across a wide range of data characteristics. For example, LSTA is higher than the most competitive one by a large margin in several challenging scenarios, including background clutter, deformation, fast motion, motion blur, and occlusion. This provides solid evidence of the strong robustness of our model.

%----------------------------------- Qualitative results: DAVIS2016, YouTube-objects, FBMS  ---------------------------------------
\begin{figure}[!t]
	\centering
	\includegraphics[width=0.7\textwidth]{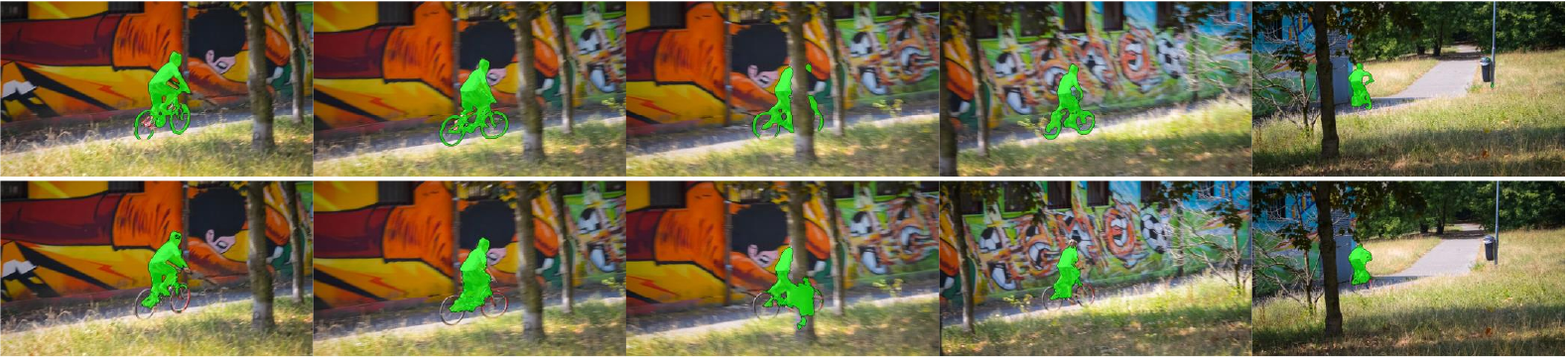}
	\includegraphics[width=0.7\textwidth]{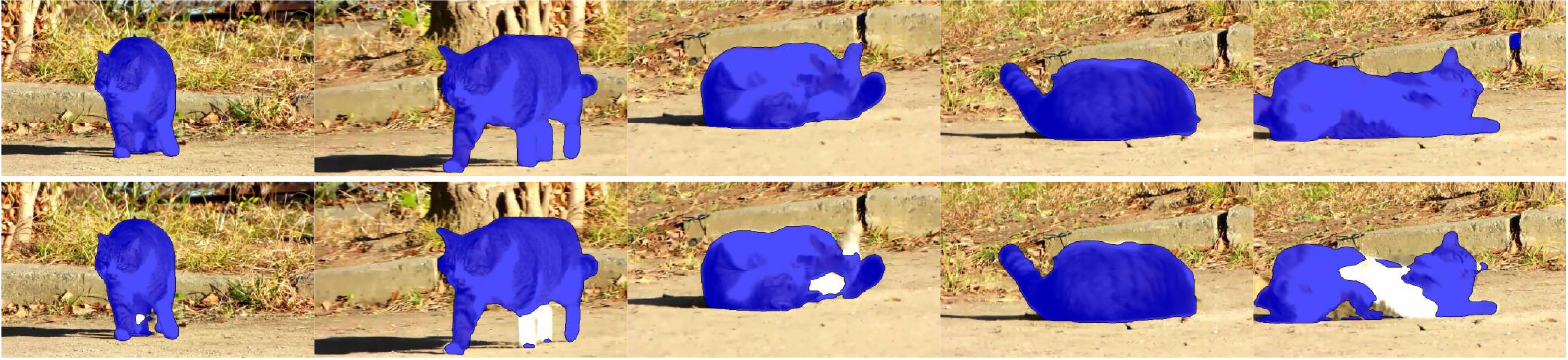}
	\includegraphics[width=0.7\textwidth]{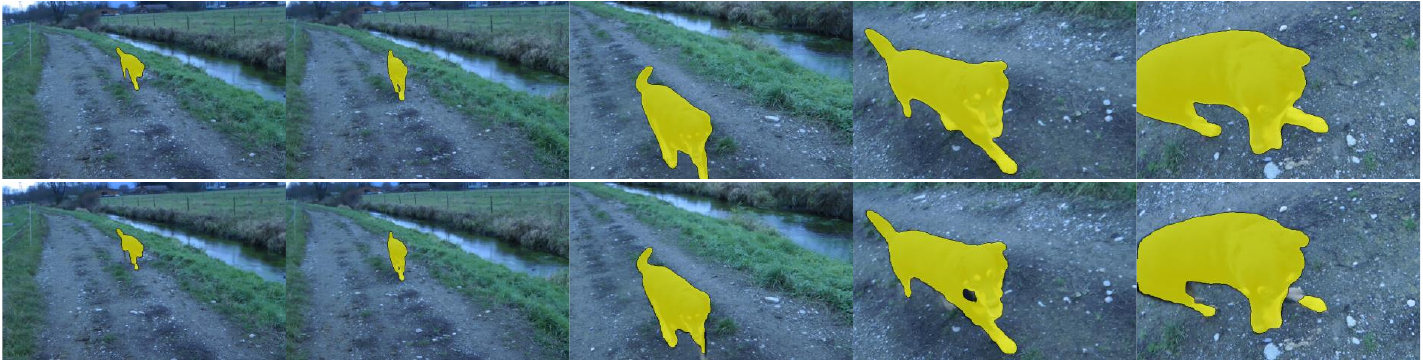}
	\caption{Segmentation results of LSTA (Row~1, 3, 5) and MATNet (row~2, 4, 6) on three randomly selected video from DAVIS2016 (row~1\&2), YouTube-objects (row~3\&4), and FBMS (row~5\&6) data sets, respectively.}
	\label{fig:res_vis_dv}
\end{figure}

\subsection{Qualitative Results}
\begin{figure}[!t]
	\centering
	\includegraphics[width=0.7\textwidth]{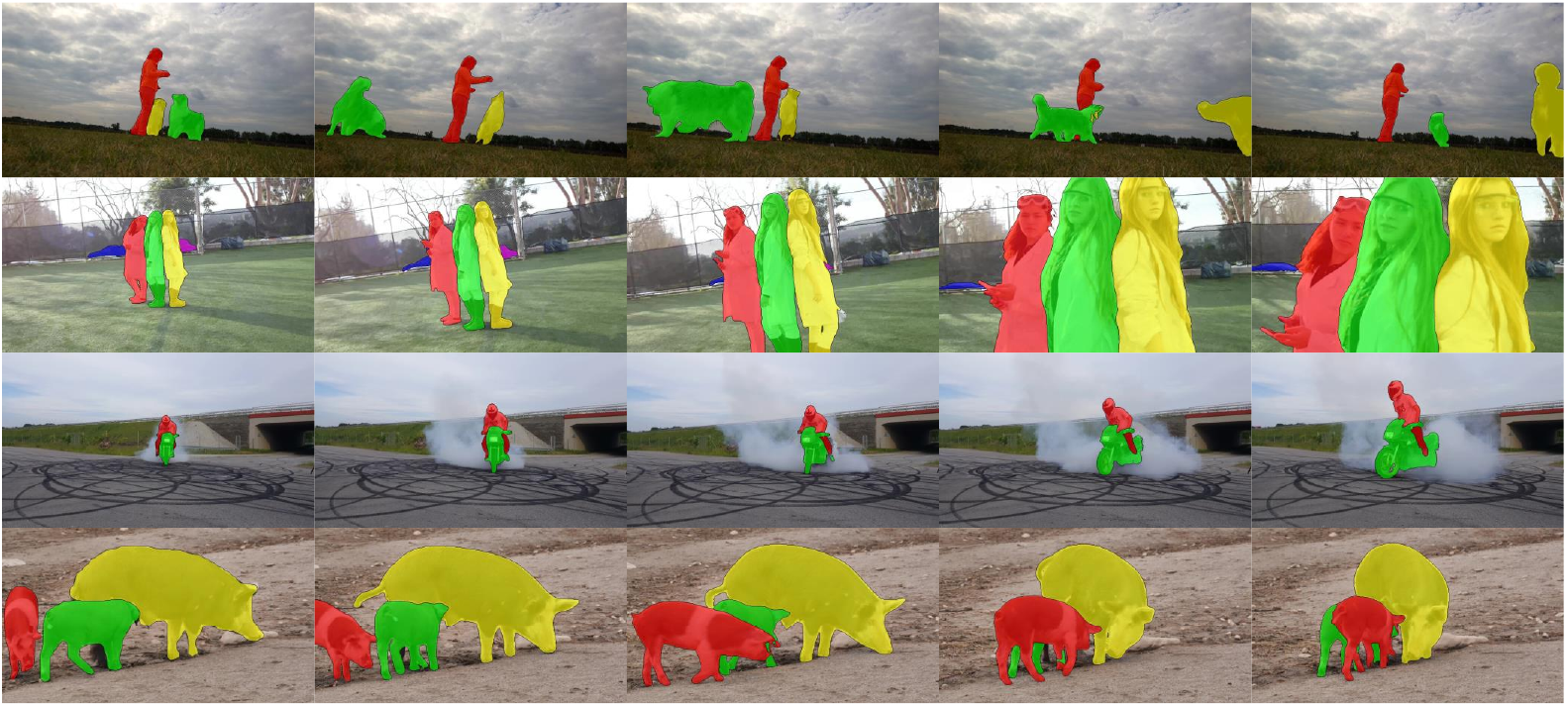}
	\caption{Segmentation results of multiple objects by LSTA on DAVIS2017.}
	\label{fig:davis17multi}
\end{figure}

To give an intuitive view on the superiority of our model, we visualize segmentation results of randomly selected video frames from DAVIS2016 \cite{perazzi-cvpr2016-davis16}, YouTube-objects \cite{prest-cvpr2012-yto}, and FBMS \cite{ochs-pami2014-fbms} in Fig.~\ref{fig:res_vis_dv}. As drawn in the figure, our model can give an accurate mask of the human-bicycle object regardless of the occluded tree (row~1), and successfully identify the local parts of objects, \eg, the feet of animals (cats in row~3 and dogs in row~5). This demonstrates that our model can well capture the moving object and the local pattern of object. In the meantime, Fig.~\ref{fig:davis17multi} shows the masks of multiple objects in four video sequences randomly chosen from DAVIS2017, and the results validate that LSTA enjoys satisfying discriminative ability of distinct categories in the same scenario.

\begin{figure}[!t]
	\centering
	\includegraphics[width=0.7\textwidth]{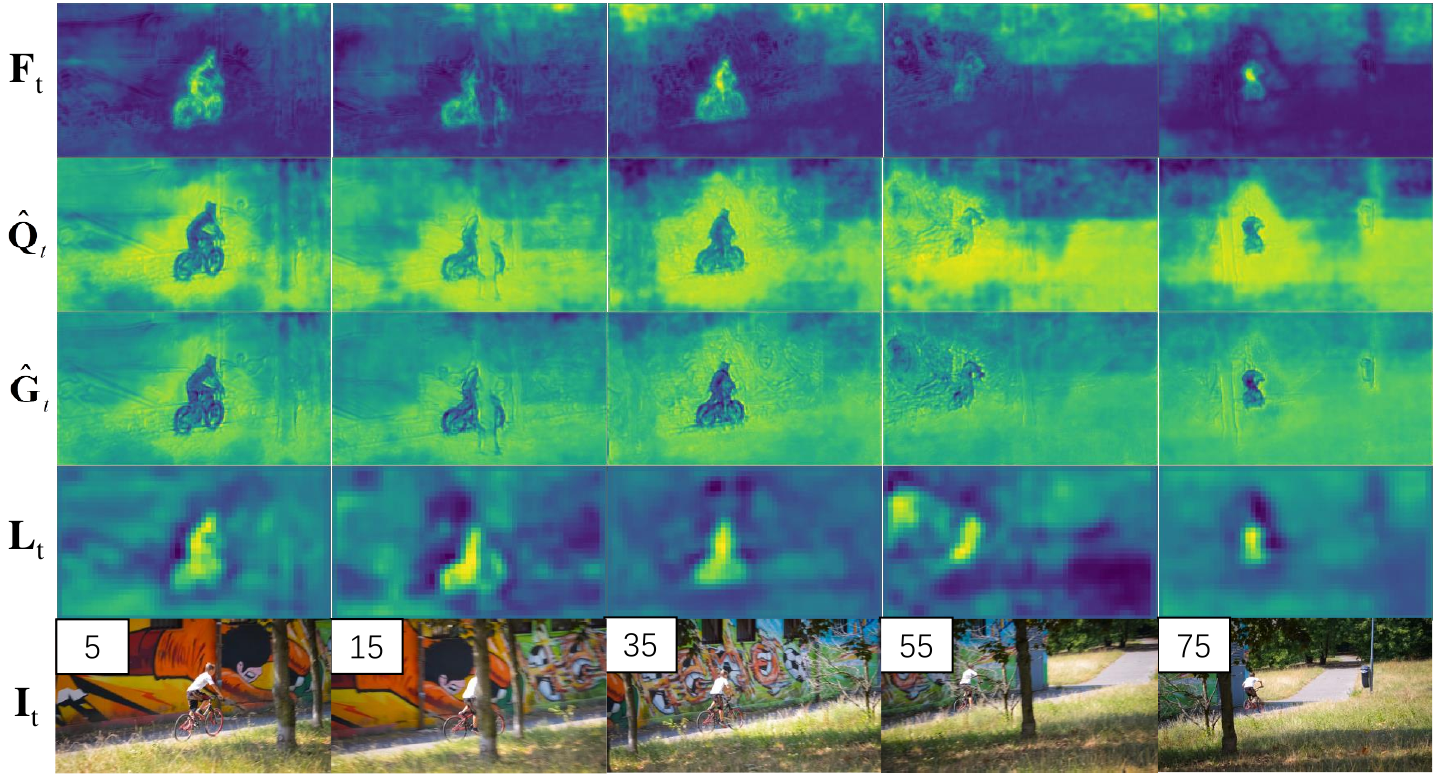}
	\caption{Feature map visualization of LSTA on DAVIS2016.}
	\label{fig:feauremap}
\end{figure}

Moreover, we exhibit the visualized feature maps generated during the intermediate procedures of inference on a video from DAVIS2016 in Fig.~\ref{fig:feauremap}. The first row shows the appearance feature map after Encoder, the second row shows the enhanced appearance feature map after Conv2D, the third and the fourth rows show the global and the local feature maps by passing LTM and STA, respectively. As vividly illustrated in these images, global feature maps are good at encoding the object contours, and local feature maps indeed play a similar role of optical flow in discovering the pattern of object motion.

\begin{figure}[!t]
	\centering
	\includegraphics[width=0.68\textwidth]{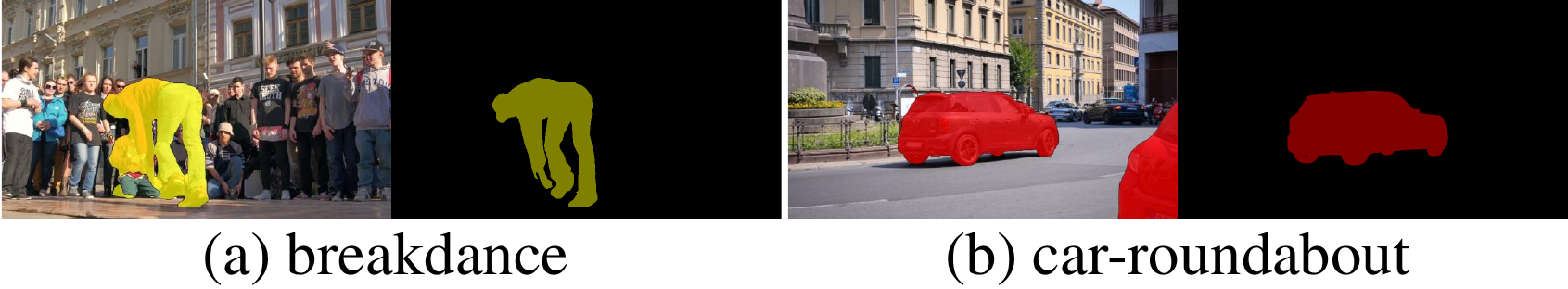}
	\caption{Failure cases on DAVIS2016. The former is the predicted mask by LSTA, the latter is the ground-truth mask.}
	\label{fig:failurecase}
\end{figure}

In addition, we show some failure cases of our model on DAVIS2016 in Fig.~\ref{fig:failurecase}. For primary objects, such as a dancing man in case (a) and a running car in case (b), there are noisy objects with similar appearance, \eg, the boy closest to the dancing man and another car in the right corner, which prevents our model from accurately segmenting the primary objects. This might be because our model is dependent of visual appearance, so one can resort to instance segmentation to alleviate the problem of similarity interference in appearance.

%----------------------------------- Qualitative results: DAVIS2017  ---------------------------------------
%

%-------------------------- Qualitative results: feature maps on DAVIS2016 ------------------------------------

%-------------------------- Qualitative results: failure cases on DAVIS2016 ------------------------------------
%

%------------------------------------------------------------------------
\section{Conclusion}
\label{conclusion}
We have developed an end-to-end real-time unsupervised video object segmentation approach, named LSTA. It includes two primary blocks, \ie, Long Temporal Memory and Short Temporal Attention, which encode both long-range and short-range spatiotemporal pixel relations of the current frame and the past frames, respectively. The former LTM captures those constantly present objects from a global view, while the latter STA models the pattern of moving objects from a local view. Moreover, we have explored the performance of our method on several benchmark datasets. Both quantitative records and qualitative visualization results indicate the superiority of the proposed approach, including more promising segmentation masks, real-time inference speed, and robustness to some deformations or occlusions. 

There still exist some limitations in our LSTA method to be addressed in the future. 1) Due to the lacking of supervision frame, it is difficult to capture those small or tiny objects, and which can be solved by designing unsupervised VOS methods tailored for small objects. 2) It fails to handle the objects with occlusions, which often appear in real-world applications. Hence, a heuristic way is to adopt the video inpainting method to recover the occluded parts first. 3) Compared to the current vanilla method, it will be interesting to consider using additional knowledge, such as referring expressions and object detections, so as to further improve the performance in unsupervised setting.

\section*{Acknowledgment}
This work was supported in part by Zhejiang Provincial Natural Science Foundation of China under Grants LR23F020002, LY22F020012, in part by ``Pioneer" and ``Leading Goose" R\&D Program of Zhejiang, China under Grants 2023C01221, 2022C03132.

\small
%\bibliography{lsta_uvos}

\end{document}